\documentclass[journal]{IEEEtran}

\ifCLASSINFOpdf
\else
\fi

\usepackage{amsmath,amssymb,amsfonts}
\usepackage{algorithmic}
\usepackage{graphicx}
\usepackage{float}
\usepackage{textcomp}
\usepackage{xcolor}
\usepackage{amsfonts,amssymb}
\usepackage{verbatim} 
\usepackage{stfloats}
\usepackage{multirow}
\usepackage{microtype}
\usepackage{optidef}
\usepackage{hyperref} 
\usepackage{svg}
\usepackage{color}

\usepackage{bm} 
\usepackage{stfloats} 

\usepackage{multirow}
\usepackage{booktabs}
\usepackage{bigstrut}
\usepackage{hyperref}

\usepackage{threeparttable} 
\usepackage{makecell} 
\usepackage{algorithm2e} 
\usepackage{adjustbox}

\usepackage[english]{babel}
\usepackage{amsthm}

\newtheorem{proposition}{Proposition}
\allowdisplaybreaks
\usepackage{caption}
\usepackage{subcaption}

\usepackage[table]{colortbl}

\newfloat{figtab}{htb}{fgtb}
\makeatletter
  \newcommand\figcaption{\def\@captype{figure}\caption}
  \newcommand\tabcaption{\def\@captype{table}\caption}
\makeatother

\begin{document}
\title{

Control-Aware Trajectory Predictions for Communication-Efficient Drone Swarm Coordination in Cluttered Environments
}

\author{Longhao Yan, Jingyuan Zhou, and Kaidi Yang

\thanks{The authors are with the Department of Civil and Environmental Engineering, National University of Singapore, Singapore 119077. Email:{\{longhao.yan, jingyuanzhou\}@u.nus.edu, 
 kaidi.yang@nus.edu.sg}. \textit{Corresponding author}: Kaidi Yang. }
 \thanks{This research was supported by the Temasek Laboratory (TL)@NUS under its Seed Grant (A-0003235-48-00) and by the Singapore Ministry of Education (MOE) under NUS Start-Up Grant (A-8000404-01-00).}
 \thanks{This work has been submitted to the IEEE for possible publication. Copyright may be transferred without notice, after which this version may no longer be accessible.}
}

\maketitle

\begin{abstract} 
Swarms of Unmanned Aerial Vehicles (UAV) have demonstrated enormous potential in many industrial and commercial applications. However, before deploying UAVs in the real world, it is essential to ensure they can operate safely in complex environments, especially with limited communication capabilities. To address this challenge, we propose a control-aware learning-based trajectory prediction algorithm that can enable communication-efficient UAV swarm control in a cluttered environment. Specifically, our proposed algorithm can enable each UAV to predict the planned trajectories of its neighbors in scenarios with various levels of communication capabilities. The predicted planned trajectories will serve as input to a distributed model predictive control (DMPC) approach. The proposed algorithm combines (1) a trajectory prediction model based on EvolveGCN, a Graph Convolutional Network (GCN) that can handle dynamic graphs, which is further enhanced by compressed messages from adjacent UAVs, and (2) a KKT-informed training approach that applies the Karush-Kuhn-Tucker (KKT) conditions in the training process to encode DMPC information into the trained neural network. We evaluate our proposed algorithm in a funnel-like environment. Results show that the proposed algorithm outperforms state-of-the-art benchmarks, providing close-to-optimal control performance and robustness to limited communication capabilities and measurement noises. 
\end{abstract}


\IEEEpeerreviewmaketitle

\section{Introduction}
\IEEEPARstart{A}{} swarm of Unmanned Aerial Vehicles (UAV) refers to a set of aerial robots, such as drones, that fly in a flock to achieve a certain goal. UAV swarms have demonstrated enormous potential in many industrial and commercial applications, such as search and rescue, remote sensing, and construction~\cite{chung2018survey,mcguire2019minimal,zhou2022swarm,yel2022meta,musil2022spheremap,wang2023bearing,fan2024air}. 
Before the massive deployment of UAV swarms in the real world, it is essential to ensure they can operate safely in complex environments such as cluttered environments where static obstacles and moving agents fill the space in an unorganized manner. 

It has been widely demonstrated that decentralized control can coordinate UAV swarms in a safe and efficient manner, thanks to its benefits in better scalability and reliability~\cite{yu2020composite,ho2020decentralized,zhou2023racer,liu2023decentralized,huang20242copre}. One category of decentralized control approaches is based on potential fields (PF) that make fast reactive decisions with behavioral rules inspired by biological swarms (e.g., birds and fish) \cite{vasarhelyi2018optimized,de2022insect,petravcek2020bio,soria2021predictive}. Although PF-based approaches have been demonstrated to be efficient, they rarely consider the physical characteristics of UAVs and can yield infeasible or overly conservative decisions, which may not be suitable for cluttered environments with dense obstacles. 
An alternative category of approaches is based on optimization, e.g., Distributed Model Predictive Control (DMPC), whereby each UAV solves a local optimization model in a decentralized manner to determine its trajectory  \cite{soria2021distributed,tordesillas2021mader,wang2022geometrically,quan2022distributed}. Compared to PF-based approaches, optimization-based approaches can explicitly account for physical constraints and generate real-time, collision-free trajectories for UAV swarms in cluttered environments. However, optimization-based approaches typically require UAVs to share planned trajectories with their neighbors at a high frequency to avoid collisions, which heavily relies on reliable communication and may not be robust in scenarios with limited communication capabilities. 

Recently, several works have attempted to develop communication-efficient control strategies for UAV swarms and other decentralized multi-robot systems, aiming to reduce communication costs for exchanging state information crucial for decision-making and control \cite{zhu2021learning,guo2021uav,shiri2020communication,sohrabi2022learning,mangalam2021goals}. 
Existing research can be divided into two categories. The first category predicts planned trajectories based on historical states measured by onboard sensors~\cite{liu2019trajectory,zhang2022ai,wang2021learning,li2023interaction,sebastian2023lemurs}. For example, Ref. \cite{zhu2021learning} developed a learning-based approach based on recurrent neural networks to predict the planned trajectories of each UAV, which does not require communications. However, this method overlooks the frequently changing topology caused by the complex interactions among UAVs in cluttered environments.
The second category seeks to compress the communicated messages on planned trajectories to reduce the bandwidth requirement~\cite{zhao2021gridless,wu2022dynamic,dong2022mr}. For example, Ref. \cite{guo2021uav} proposed a UAV path discretization and compression algorithm that simultaneously optimizes the selection of waypoints and the subpath trajectories while ensuring the optimal performance of trajectory optimization. However, the proposed algorithm only focused on each individual path without exploiting their correlations. Ref. \cite{shiri2020communication} developed a neural network-based mean-field approximation approach to approximate the solution to the Hamilton-Jacobi-Bellman (HJB) and Fokker-Planck-Kolmogorov (FPK) equations, which can effectively minimize communication. To avoid violating required conditions, this work introduced Federated Learning to enable the sharing of model parameters between UAVs. However, this work requires the online training and sharing of the neural network parameters, which even at low communication frequency, can be a large amount of data to exchange. 

From the literature mentioned above, we have identified two research gaps in existing research on communication-efficient UAV swarm coordination. First, existing works either attempt to predict trajectories from historical states or compress communicated messages on planned trajectories but rarely combine both approaches such that these two types of information can compensate for each other. Moreover, many existing works overlook the changing topology of UAV swarms or have made strong assumptions of reliable communications and have yet to test their proposed algorithms in scenarios with cluttered environments. 
Second, existing works rarely consider information about the decision-making process of the UAV swarm, e.g., optimization models, which can be common knowledge for all UAVs in the same swarm. Integrating such knowledge can help ensure that the predicted trajectories are more physically meaningful. 

In this paper, we address the aforementioned research gaps by developing a learning-based algorithm that enables each UAV to predict the trajectories of its neighbors in cluttered environments with various levels of communication capabilities. 
Our algorithm combines three types of information: (i) the historical states of each UAV and its neighbors, (ii) compressed messages communicated from neighbors, if available, and (iii) the information about the decision-making process of each UAV, characterized by a convex optimization-based DMPC approach. Our algorithm combines the first two types of information by integrating two modules in a Bayesian framework: a graph convolutional network (GCN)-based trajectory prediction module and a Variational Auto-Encoder (VAE)-based trajectory compression and reconstruction module. 
Our trajectory prediction module can handle dynamic scenarios with time-varying swarm alignment, where the neighbors of a given UAV constantly change. This is achieved by leveraging an EvolveGCN architecture~\cite{pareja2020evolvegcn} that can track the change in the underlying graph. The third type of information, i.e., the information of the convex optimization model upon which each UAV makes decisions, is encoded in a Karush–Kuhn–Tucker (KKT) conditions-informed training method \cite{amos2017optnet} that trains the two modules together. 

\emph{Statement of Contribution}. The main contributions of this paper are two-fold. First, we devise a learning-based algorithm to predict the trajectories of surrounding UAVs in cluttered environments with various levels of communication capabilities  (bandwidths, packet loss rates, lower frequencies, etc.), which is achieved by combining an EvolveGCN-based trajectory prediction model with a VAE-based trajectory compression and reconstruction model in a Bayesian framework. Second, we train the developed neural network via a KKT-informed approach that enables the trajectory prediction algorithm to encode the information about the decision-making process of UAVs to improve control performance. To the best of our knowledge, no existing trajectory prediction algorithms have explicitly encoded knowledge about the decision-making processes of the agents. 

This paper is organized as follows. Section \ref{sec:problem statement} describes the problem statement, Section \ref{sec:methodology} presents our methodology, Section~\ref{sec:simulation} presents the simulation settings and results, and Section~\ref{sec:conclusion} concludes the paper. 

\section{Problem Statement} \label{sec:problem statement}

\begin{figure}[htbp]
    \centering
    \includegraphics[width=0.3\textwidth]{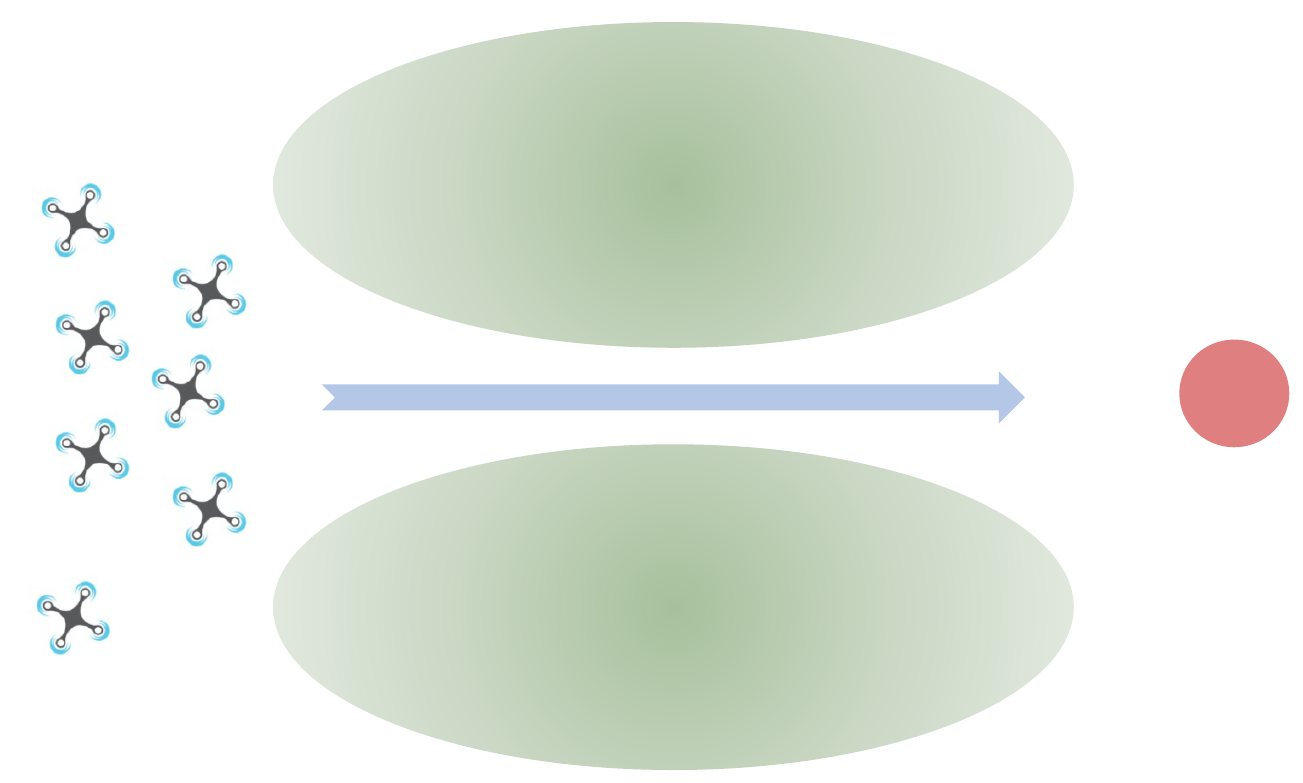}
    \caption{Illustration of the considered scenario, where the red point represents the goal of the swarm, and the green ellipsoids represent the obstacles. }
    \label{fig:scenario}
\end{figure}
As shown in Figure~\ref{fig:scenario}, we consider a swarm of UAVs indexed by $i \in \mathcal{V}$ in a funnel-like environment where the obstacles, indexed by $m\in\mathcal{M}$, are convex and modeled as 3-dimensional ellipsoids, each parameterized by a center point $\bm{C}_m\in\mathbb{R}^3$ and an affine transformation matrix $\bm{E}_m\in\mathbb{R}^{3\times 3}$. 
The goal of the UAV swarm is to safely navigate to a predefined migration point (represented by the red point in Figure \ref{fig:scenario}).
The considered time horizon is discretized as a set of intervals $\mathcal{K}=\{1,2,\cdots, K\}$ of a given size $\Delta T$ indexed by $k$. The state of UAV $i\in \mathcal{V}=\{1,2,\cdots,n\}$ at time step $k\in \mathcal{K}$ is represented by $\bm{x}_i(k)=\left(\bm{p}_i(k), \bm{v}_i(k)\right)\in\mathbb{R}^6$, where $\bm{p}_i(k)\in\mathbb{R}^3$ and $\bm{v}_i(k)\in\mathbb{R}^3$ denote its location and velocity at time step $k$, respectively. The dynamics of UAV $i$ can be modeled as a discrete linear system with an underlying position controller: 
\begin{align}
    \bm{x}_i(k+1) = \bm{A}_i\bm{x}_i(k)+\bm{B}_i\bm{u}_i(k) \label{eq:dynamics}
\end{align}
where the control action $\bm{u}_i(k)$ represents the desired position at time step $k$, and $\bm{A}_i$ and $\bm{B}_i$ are constant matrices.

Each UAV $i$ can collect two types of information: (1) the real-time states of other UAVs within a sensing range, denoted by $\bm{x}_j(t)$ for the states of UAV $j$ at time step $t$, and (2) the messages sent by other UAVs within a communication range, denoted by $\bm{z}_j(t)$ for the message sent by UAV $j$ at time step $t$. 
For sensing, we assume each UAV can fully sense the entire swarm, so the sensing graph $\mathcal{G}_{\text{sen}} = (\mathcal{V}_{\text{sen}}, \mathcal{E}_{\text{sen}})$ is a complete graph. This can be achieved through the multi-sensor fusion of LiDAR, cameras, ultra-wideband, etc., which has been demonstrated effective even in cluttered environments~\cite{gadde2021fast,han2021fast}.
For communication, each UAV is assumed to be able to communicate with the nearest up to four UAVs within a given range, forming a communication graph $\mathcal{G}_{\text{comm}} = (\mathcal{V}_{\text{comm}}, \mathcal{E}_{\text{comm}})$. Let neighbors $\mathcal{N}_i$ denote the set of UAVs that UAV $i$ can communicate with. 
We make the following remarks regarding the communication graph $\mathcal{G}_{\text{comm}}$. First, we consider the nearest up to four neighbors because the movement of the UAV is mainly influenced by these immediate neighbors. Such a treatment helps balance the swarm performance, computational complexity of the controller, and communication burden~\cite{soria2021distributed}.
Second, we consider the communication graph $\mathcal{G}_{\text{comm}}$ to be dynamic, with both the node set $\mathcal{V}_{\text{comm}}$ and edge set $\mathcal{E}_{\text{comm}}$ varying over time to represent changing neighbors.  
Such an assumption is realistic since the alignment of UAVs can change dynamically, especially in cluttered environments. 
Third, the messages sent by UAV $j$ to UAV $i$ can contain possibly compressed information about UAV $j$'s planned trajectories, which will be used to facilitate the decision-making of UAV $i$. Note that the trajectory information is compressed to cater to the limited bandwidth of the communication channel. To simplify the implementation, we assume that each UAV sends identical structured messages to all its neighbors on the communication graph. Nevertheless, this assumption can be easily extended such that UAVs can share customized messages with their neighbors under specific conditions. 

The decision-making process of each UAV follows the quadratic programming (QP)-based DMPC approach proposed in \cite{soria2021distributed}. Note that this DMPC-based control algorithm is chosen thanks to its simplicity of implementation. Our proposed trajectory prediction algorithm can also be used with other convex optimization-based decision algorithms.  

In the DMPC settings, each UAV $i\in\mathcal{V}$ solves an embedded quadratic programming (QP) problem at each time step $k$ to obtain its decision, i.e., the planned trajectory within the next $P$ time steps denoted as $\bm{U}_i^k = \{\bm{u}_i(k+h|k)\}_{h=0}^{P-1}$. 
For each UAV, the input to the QP problem involves (1) its current location and speed, (2) the real-time location information about the obstacles and the UAVs within its neighborhood, and (3) the predicted trajectory of its neighbors in the planning horizon, denoted by $\tilde{\bm{U}}_{-i}^{k}$. 

To better formulate collision-avoidance constraints, the planned trajectory of UAV $i$, $\bm{U}_i^k$, is parameterized by $l$ 3-dimensional Bézier curves of order $d$, whereby each of these Bézier curves can be uniquely characterized by a set of $d+1$ 3-dimensional control points. Consequently, the planned trajectory $\bm{U}_i^k$ is characterized by a vector $\bm{w}_i^k\in\mathbb{R}^{3l(d+1)}$ formed by $l(d+1)$ 3-dimensional control points. This parameterization allows us to define continuous trajectories in a polynomial form $f(\bm{w}_i^k)$. The relation between the planned trajectory and the control points can be described as $\bm{U}_i^k = \bm{F}_i\bm{w}_i^k$, where $\bm{F}_i$ indicates the coefficients of the Bézier curves.
 
With the aforementioned trajectory parametrization, the embedded QP problem of the DMPC scheme for UAV $i$ can be described in Eq.(\ref{eq:obj})-(\ref{eq:nonneg}), as inspired by \cite{soria2021distributed}. 
\begin{subequations}    
\begin{align}        \min_{\bm{w}_i^k,\bm{\zeta}_i^k,\bm{\epsilon}_i^k,\bm{\delta}_i^k} 
    &~\sum_{\tau=1}^{P} q_{\text {mig }}\left\|\boldsymbol{p}_{i}(k+\tau \mid k)-\boldsymbol{p}_{\text {mig }}\right\|_{2}^{2}  \notag \\
    &+ l_{\mathrm{saf}} \boldsymbol{\zeta}_{i m}\left(k_{\mathrm{coll}, i} \mid k\right)+q_{\mathrm{saf}} \boldsymbol{\zeta}_{i m}^{2}\left(k_{\mathrm{coll}, i} \mid k\right)  \notag \\
    &+\sum_{j \in \mathcal{N}_{i}} \sum_{\tau=0}^{P-1}\left(l_{\text {saf }} \boldsymbol{\epsilon}_{i j}(k+\tau \mid k)+q_{\text {saf }} \boldsymbol{\epsilon}_{i j}^{2}(k+\tau \mid k)\right)  \notag \\
    &+ \sum_{j \in \mathcal{N}_{i}} \sum_{\tau=0}^{P-1}\left(l_{\text {coh }} \boldsymbol{\delta}_{i j}(k+\tau \mid k)+q_{\text {coh }} \boldsymbol{\delta}_{i j}^{2}(k+\tau \mid k)\right) \notag \\
    &+ \sum_{\tau=0}^{P-1} q_{\text {eft }}\left\|\frac{d^{2}}{d t^{2}} f(\bm{w}_i)(k+\tau \mid k)\right\|_{2}^{2}  \label{eq:obj} \\ 
     \text{s.t. } & {\boldsymbol{A}_{\text{dyn},i}} \boldsymbol{w}^k_{i} \leq \boldsymbol{b}_{\text{dyn},i} \label{eq:dyn}\\ 
    & \boldsymbol{A}_{\text{cont},i}  \boldsymbol{w}^k_{i} = \boldsymbol{b}_{\text{cont},i} \label{eq:cont}\\ 
    & {\boldsymbol{A}^k_{\text{saf-agent},i}}\left(\tilde{\bm{U}}_{-i}^{k}\right) 
    \left[({\boldsymbol{w}}^k_{i})^T, (\boldsymbol{\epsilon}^k_{i})^T\right]^T \notag \\ 
    & \quad\quad\leq \boldsymbol{b}^k_{\text{saf-agent},i}\left(\tilde{\bm{U}}_{-i}^{k}\right) \label{eq:saf-agent} \\
    & {\boldsymbol{A}^k_{\text{coh},i}}\left(\tilde{\bm{U}}_{-i}^{k}\right) \left[(\boldsymbol{w}^k_{i})^T, (\boldsymbol{\delta }^k_{i})^T\right]^T \leq \boldsymbol{b}^k_{\text{coh},i}\left(\tilde{\bm{U}}_{-i}^{k}\right) \label{eq:coh} \\
    & {\boldsymbol{A}^k_{\text{saf-obs},i}} [({\boldsymbol{w}}^k_{i})^T, (\boldsymbol{\zeta}^k_{i})^T]^T \leq \boldsymbol{b}^k_{\text{saf-obs},i} \label{eq:saf-obs}\\
    & \bm{\epsilon}_i^k,\bm{\delta}_i^k,\bm{\zeta}_i^k \geq 0 \label{eq:nonneg}
\end{align}
\end{subequations}
where $\boldsymbol{p}_{\text {mig }}$ is the pre-defined goal position that UAV swarm aims to reach,  $k_{\mathrm{coll}, i}$ is the time that UAV $i$ would first collide with obstacle $m$ in its predicted horizon if the control actions remain the same. 
Slack variables $\boldsymbol{\zeta}, \boldsymbol{\epsilon}, \boldsymbol{\delta}$ represent the violation of the safe agent-obstacle distance, safe inter-agent distance, and desired inter-agent cohesion distance, respectively.
The objective function is a quadratic function involving (i) migration cost that penalizes the distance to a pre-defined migration point, (ii) agent-obstacle and inter-agent collision costs penalizing the UAV of getting too close to obstacles and its neighbors, (iii) cohesion cost to ensure that adjacent UAVs stay closer than a threshold cohesion distance, and (iv) control effort cost, characterized by the second derivative of the polynomial Bézier curve (i.e., the acceleration), which is minimized to reduce the energy spent on executing the control action.
Constraint (\ref{eq:dyn}) ensures the feasibility of the dynamics given by the planned trajectory parameterized as the Bézier curve. Constraint (\ref{eq:cont}) imposes $C^2$-continuity requirement on the derived Bézier curve. Constraints (\ref{eq:saf-agent}) and (\ref{eq:coh}), respectively, indicate the collision avoidance and cohesion between UAV $i$ and its surrounding UAVs to bound the distance between them. 
Constraint (\ref{eq:saf-obs}) indicates the collision avoidance between UAV $i$ and obstacles. Note that constraints (\ref{eq:saf-agent})-(\ref{eq:saf-obs}) are linearized from nonlinear constraints using first-order Taylor expansion. The constraint (\ref{eq:nonneg}) requires these relaxation terms to be nonnegative. 

Further note that constraints (\ref{eq:saf-agent}) and (\ref{eq:coh}) depend on the predicted trajectories of the surrounding UAVs, i.e., $\tilde{\bm{U}}_{-i}^{k}$. Hence, the quality of decisions, especially on inter-agent safety and cohesion, heavily depends on the accuracy of such a prediction. Originally in \cite{soria2021distributed}, such predictions are assumed to be planned trajectories always available with moderate noises, as they can be obtained through communicating with surrounding UAVs at each time step. However, since the communication channel may have limited bandwidth and frequency, as well as possibility of packet loss, it may not be feasible to share the entire trajectory at each time step. Therefore, it is important to develop a communication-efficient and robust method for trajectory prediction, which will be developed in Section~\ref{sec:methodology}.  

\section{Control-aware learning-based trajectory prediction} \label{sec:methodology}

\begin{figure*}[htbp]
    \centering
    \includegraphics[width=0.99\textwidth]{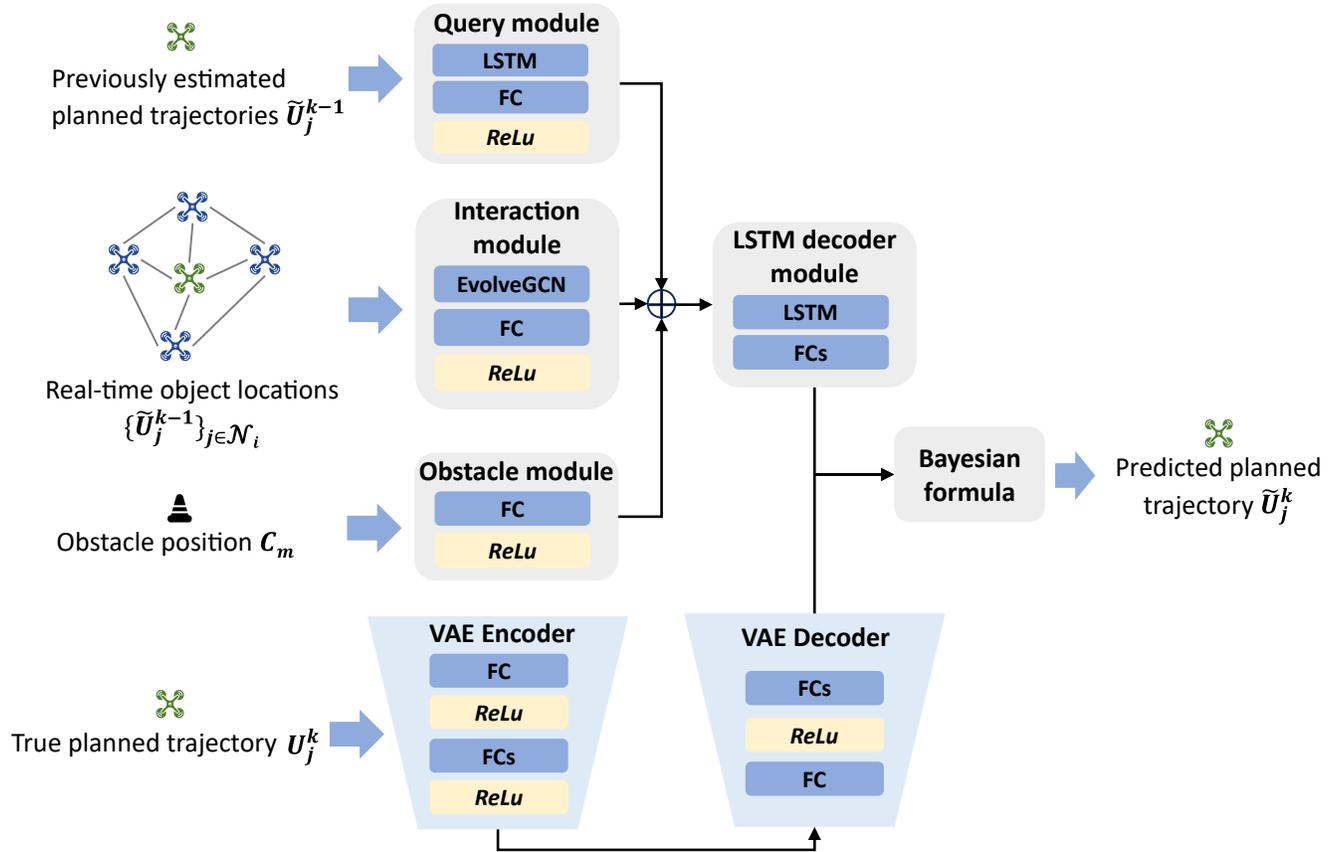}
    \caption{Framework of the trajectory prediction algorithm}
    \label{fig:method_predictor}
\end{figure*}
\begin{figure*}[htbp]
    \centering
    \includegraphics[width=0.95\textwidth]{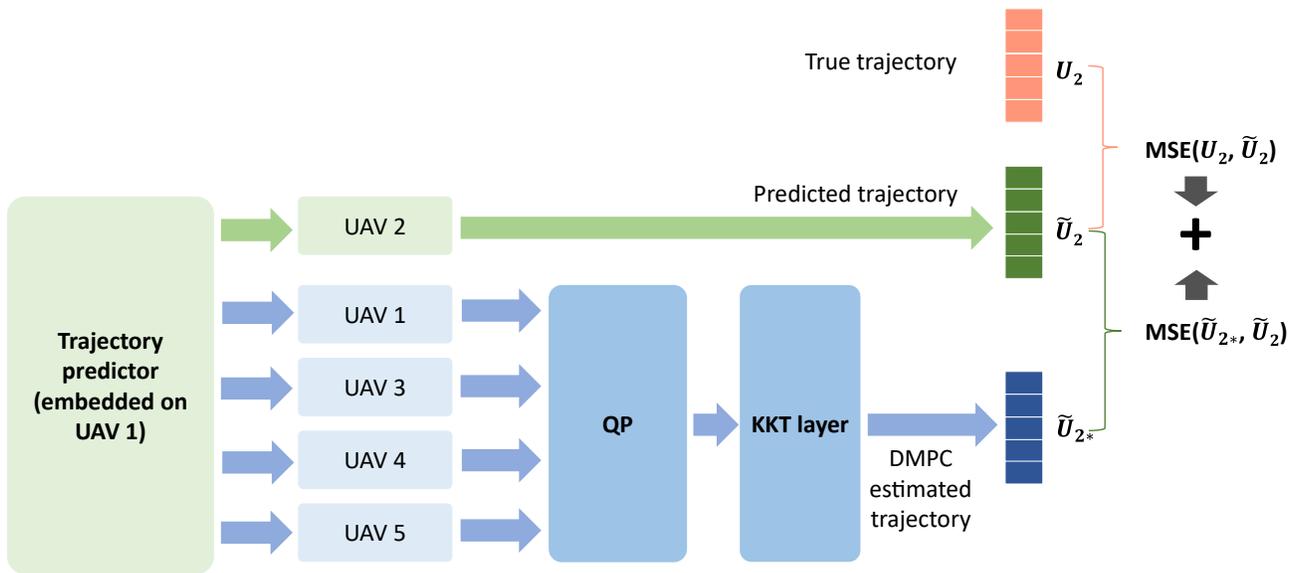}
    \caption{Example of KKT-informed training process, whereby UAV 1 wants to estimate the planned trajectories of UAV 2. UAV 1, 3, 4, and 5 are the closest neighbors of UAV 2.
    }
    \label{fig:method_KKT_training}
\end{figure*}

To address the challenge imposed by limited communication, we devise a learning-based approach to predict the trajectories of the surrounding UAVs of each UAV $i\in\mathcal{V}$. 
Here, we seek to predict the trajectories of surrounding UAVs by estimating their planned trajectories at the current time step. 
The reason is three-fold. First, due to embedded motion control modules in UAVs, we envision the planned trajectory to be close to the actual trajectory, especially if the planning horizon is relatively short. Second, the estimation of the planned trajectories enables us to encode more physical information (e.g., the decision-making process of UAVs) and thus can be more sample-efficient in training. Third, using the planned trajectories to represent future trajectories is commonly seen in existing literature \cite{luis2020online, soria2021distributed, zhu2021learning, sebastian2023lemurs, brandstatter2023multi}, which enables us to better integrate our approach with state-of-the-art control algorithms. 
Mathematically, let $\tilde{\bm{U}}_{ij}^{k}$ be the predicted trajectory of UAV $j$ calculated by UAV $i$ at the beginning of time step $k$. For brevity, we simplify the notion $\tilde{\bm{U}}_{ij}^{k}$ to $\tilde{\bm{U}}_{j}^{k}$ by omitting $i$ if there is no ambiguity. 
Therefore, the predicted trajectory of all neighbors of the UAV $i$, $\tilde{\bm{U}}_{-i}^{k}$, can be represented by $ \tilde{\bm{U}}_{-i}^{k} = \{\tilde{\bm{U}}_{j}^{k}\}_{j \in \mathcal{N}_i}$.

The proposed trajectory prediction framework involves a trajectory predictor illustrated in Figure \ref{fig:method_predictor} and a KKT-informed training approach illustrated in Figure \ref{fig:method_KKT_training}. The trajectory predictor combines the observations made by the ego UAV and the compressed messages shared by the target UAV, if any. The KKT-informed training approach facilitates training by devising a KKT-based differential layer to encode the structural information of the embedded QP into the trained neural network. 
The details of these components are detailed below in Sections~\ref{sec:EG}--\ref{sec:KKT}. Note that the developed learning-based model is implemented in each UAV with identical parameters.

\subsection{Trajectory predictor} \label{sec:EG}
We consider the scenario where UAV $i$ predicts the trajectory of UAV $j$ using its own observations and the messages shared by UAV $j$, if any. 
To this end, our trajectory prediction algorithm uses a Bayesian framework to combine the output of two modules: (i) an EvolveGCN that processes the observations of the ego UAV and yields predicted trajectories (denoted by $\tilde{\bm{U}}_{j,\text{EG}}^{k}$) and (ii) a VAE, whereby an encoder implemented on UAV $j$ compresses the planned trajectory of UAV $j$, and a decoder implemented on UAV $i$ reconstructs the compressed message to obtain the reconstructed trajectories (denoted by $\tilde{\bm{U}}_{j,\text{VAE}}^{k}$). Next, we provide the details of these two modules and the Bayesian framework.

\vspace{0.3em} \noindent \textbf{(1) EvolveGCN-based trajectory prediction module}. 
This module enables UAV~$i$ to predict the trajectories of its neighbor $j$, $\tilde{\bm{U}}_{j, \rm{EG}}^{k}$, at each time step $k$, leveraging (i) the previously predicted trajectories of UAV $\tilde{\bm{U}}_{j}^{k-1}$ at time step $k-1$, (ii) the location information within a horizon of $H$ past time steps $\bm{p}^{k,\text{hist}}=\{\bm{p}_j^{k-h}\}_{j\in\mathcal{N}_i,h\in \{0,1,\cdots,H\}}$ of surrounding UAVs, measured by the onboard sensors of UAV $i$, and (iii) the geometric information of obstacles. 

As shown in Figure~\ref{fig:method_predictor}, the EvolveGCN (EG) architecture involves four parts: 
(i) a Long Short-Term Memory (LSTM)-based query module to characterize the correlations between trajectories predicted at consecutive time steps, (ii) an EvolveGCN-based interaction module to extract the spatiotemporal correlations of UAVs on the dynamic communication graph, (iii) an obstacle encoder to extract the geometric features of obstacles, and (iv) an LSTM-based prediction module to generate estimated $\tilde{\bm{U}}_{j, \text{EG}}^{k}$ and the vector of element-wise standard deviations $\bm{\Sigma}_{j,\text{EG}}^{k}$. 
Parts (i), (iii), and (iv) build on the trajectory planning neural network developed in \cite{zhu2021learning} and extend it by allowing a probabilistic output representation for integration with VAE reconstruction outcomes. To this end, we add Part (ii) with an EvolveGCN architecture, which represents the swarm alignment by a GCN and evolves its parameters along the temporal dimension using LSTM without resorting to node embeddings. 
The details of the four parts are described as follows. 

\emph{(i) LSTM-based query module}. 
An LSTM layer with a Fully-Connected (FC) layer is leveraged to exploit the temporal correlations between the trajectories of UAV $j$ at consecutive time steps. The activation function $\pi(\cdot)$ is chosen as ReLU for its gradient stability and computational efficiency. This encoder can be mathematically represented as: 
\begin{align}
    \bm{y}_{j}^k = \text{Query}_{\text{LSTM}}\left( \tilde{\bm{U}}_{j}^{k-1}\right),~j\in\mathcal{N}_i,
\end{align}
where the output $\bm{y}_j^k$ serves as an input to the LSTM decoder, i.e., Part (iv).

\emph{(ii) EG-based interaction module}. As mentioned above, the EG layer is adopted to extract the spatiotemporal correlations of UAVs in time-varying communication graphs. The input to the EG layer includes the position information of $H$ past time steps $\bm{p}^{k,\rm{hist}}$ within the communication range of UAV $j$.

Each EG layer involves a GCN layer and an LSTM layer, whereby the GCN layer captures the topology of the communication graph of UAV $j$, represented as adjacency matrix $\bm{G}_j^k\in \mathbb{R}^{n\times n}$, where the elements involving unseen UAVs are set to zero. The LSTM layer is employed to model the evolution of the GCN weight matrix as a dynamical system. 
Specifically, let $\bm{W}^{k,(l)}$ denote the dynamic GCN weight matrix of the $l$-th layer at time step $k$, and $\bm{H}^{k,(l)}$ represents the node embeddings of the $l$-th layer at time step $k$. The EG layer can be written as:
\begin{subequations}    
\begin{align}
    \bm{H}^{k+1,(l+1)} &= \text{GCONV}(\bm{G}_j^k, \bm{H}^{k,(l)},\bm{W}^{k,(l)} ) \label{eq:GCN}\\
    \bm{W}^{k+1,(l)} &= \text{LSTM}_{\text{EG}}\left(\bm{W}^{k,(l)}\right). \label{eq:eg_lstm}
\end{align}
\end{subequations}
where the initial embedding matrix $\bm{H}^{k,(0)}$ includes the node features as the real-time locations of surrounding UAVs and the final embedding matrix $\bm{H}^{k,(L)}$ is the output of the EG layer. 
In Eq.(\ref{eq:eg_lstm}), the evolution of the weight matrix $\bm{W}^{k,(l)}$ is solely based on the weight matrix at the previous time step, which is referred to as EGCU-O. An alternative way of formulating this LSTM layer is to incorporate node embeddings $\bm{H}^{k,(l)}$, i.e., $\bm{W}^{k+1,(l)} = \text{LSTM}\left(\bm{H}^{k+1,(l)}, \bm{W}^{k,(l)}\right)$, referred to as EGCU-H \cite{pareja2020evolvegcn}. In this paper, we use EGCU-O for its simplicity. 

The output of the EG layer is further processed by a ReLu-activated FC layer to reshape the feature dimensions. The output of the interaction module, $\bm{g}^k$, serves as an input to the LSTM-based decoder, i.e., Part (iv).

\emph{(iii) FC-based obstacle module}. Since we only consider static obstacles, we design the obstacle encoder as a fully connected layer with an input of the parameters characterizing the obstacles. This encoder can be mathematically represented as: 
\begin{align}
    \bm{o}^k = \textit{ReLu}\left(\text{FC}_{\text{obs}}\left(\bm{C}_m\right)\right),~{m\in M}
\end{align}
where the output $\bm{o}^k$ serves as an input to the LSTM decoder, i.e., Part (iv).

\emph{(iv) LSTM-based decoder module}. The decoder synthesizes the outputs of the LSTM-based query module, the EG-based interaction module, and the FC-based obstacle module with an LSTM layer followed by two parallel fully connected layers to generate predicted trajectories of UAV $j$, $\tilde{\bm{U}}_{j}^{k}$ and the vector of element-wise standard deviations $\bm{\sigma}_{j,\text{EG}}^{k}$, which can be written as:
\begin{align}
    \tilde{\bm{U}}_{j,\text{EG}}^{k}, \bm{\sigma}_{j,\text{EG}}^{k} = \text{Decoder}_{\text{LSTM}}\left(\bm{y}_{j}^k, \bm{o}^k,\bm{g}^k\right)
\end{align}
where the distribution of the ground-truth trajectory is given as a Gaussian distribution with mean $\tilde{\bm{U}}_{j, \rm{EG}}^{k}$ and covariance matrix $\Sigma^{k}_{j,\text{EG}} = \left(\text{diag}(\bm{\sigma}_{j,\text{EG}}^{k})\right)^2$. We 
choose the covariance matrix as a diagonal matrix, a typical treatment in existing works to simplify the underlying neural network. 

\vspace{0.3em} \noindent \textbf{(2) VAE-based trajectory compression and reconstruction}. In communication-limited scenarios, a VAE is used for trajectory compression and reconstruction. It consists of an encoder $\phi_{\rm{e}}$ and a decoder $\phi_{\rm{d}}$. 
The encoder is leveraged by UAV $j\in\mathcal{N}_i$ to compress its planned trajectory $\bm{U}_j^{k} \in \mathbb{R}^{3P}$ into a low-dimensional message $\bm{z}_j^{k} \in \mathbb{R}^{L}$ ($L<3P$). Specifically, the encoder is implemented as a neural network that outputs a vector of means $\phi_{\rm{e}}(\bm{U}_j^{k})\in \mathbb{R}^{L}$ and standard deviations $\bm{\sigma}_{j,\text{VAE}}^{k} \in \mathbb{R}^{L}$ of independent Gaussian distributions, from which the message $\bm{z}_j^{k}$ is sampled and transmitted to UAV~$i$. After UAV $i$ receives the message $\bm{z}_j^{k}$, it uses the decoder to reconstruct the trajectory $\tilde{\bm{U}}_{j,\text{VAE}}^{k}=\phi_{\rm{d}}(\bm{z}_j^{k})$. In this paper, both the encoder network $\phi_{\rm{e}}$ and decoder network $\phi_{\rm{d}}$ are implemented as fully connected layers. In addition, the trajectory reconstructed by the decoder naturally makes VAE an estimator of the true planned trajectory.

\vspace{0.6em} \noindent \textbf{(3) Bayesian integration.} We combine the output of the EG and VAE-based modules using a simple Bayesian approach.
Specifically, we treat the outcome of the EG-based module as the prior because such prediction will always be available despite communication failure. Specifically, the prior distribution is characterized by a Gaussian distribution written as:
\begin{align}
\bm{U}_{j}^{k} | \tilde{\bm{U}}_{j,\text{EG}}^{k} \sim N\left(\tilde{\bm{U}}_{j,\text{EG}}^{k}, \Sigma^{k}_{j,\text{EG}}\right) \label{eq:prior}
\end{align}
where $\bm{U}_{j}^{k}$ represents the ground-truth  trajectories of UAV $j$ starting from time step $k$. 

If UAV $i$ receives a message $\bm{z}_j$ from UAV $j$ and reconstructs it, the reconstructed outcome, $\tilde{\bm{U}}_{j,\text{VAE}}^{k}$, will serve as an observation, i.e.,  
\begin{align}
\tilde{\bm{U}}_{j,\text{VAE}}^{k}|\bm{U}_j^{k} \sim N\left(\bm{U}_j^{k}, \Sigma^{k}_{\text{VAE}}\right),~j \in \mathcal{N}_i\label{eq:meas}
\end{align}
where the covariance matrix $\Sigma^{k}_{\text{VAE}}$ can be obtained empirically from a well-trained VAE. 

Combining Eq.(\ref{eq:prior}) and (\ref{eq:meas}), the posterior distribution can be derived using the Bayesian formula as in Eq.(\ref{eq:bayesian}). 
These outputs are combined using a Bayesian framework that calculates the following posterior 
\begin{align}
    p\left(\bm{U}_{j}^{k}\mid \tilde{\bm{U}}_{j,\text{VAE}}^{k}, \tilde{\bm{U}}_{j,\text{EG}}^{k}\right) \propto p\left(\bm{U}_{j}^{k}\mid \tilde{\bm{U}}_{j,\text{EG}}^{k}\right)p\left(\tilde{\bm{U}}_{j,\text{VAE}}^{k} | \bm{U}_{j}^{k}\right).\label{eq:bayesian}
\end{align}
where notice that $\tilde{\bm{U}}_{j,\text{VAE}}^{k}$ and $\tilde{\bm{U}}_{j,\text{EG}}^{k}$ are conditional independent. Then, the final prediction can be calculated via maximum a posteriori (MAP) as $\tilde{\bm{U}}_{j}^{k}=\arg\max_{\bm{U}_{j}^{k}} p(\bm{U}_{j}^{k}|\tilde{\bm{U}}_{j,\text{VAE}}^{k}, \tilde{\bm{U}}_{j,\text{EG}}^{k})$.

\subsection{KKT-informed training of the trajectory prediction neural network} \label{sec:KKT}
Sections~\ref{sec:EG} defines a learning-based framework for trajectory prediction with a neural network model. 
One classical method to train the neural network is to minimize the $l_2$-loss between the true planned trajectories $\bm{U}_{j}^{k}$ and the predicted trajectories $\tilde{\bm{U}}_{j}^{k}$ for each UAV $j$ at each time step $k$, written as 
\begin{align}
    \mathcal{L}_2(\tilde{\bm{U}}_{j}^{k}, \bm{U}_{j}^{k}) = ||\tilde{\bm{U}}_{j}^{k}- \bm{U}_{j}^{k}||_2^2 \label{eq:l2_loss}
\end{align}
where $ \mathcal{L}_2$ represents the $l_2$-loss function. 

Although such a classical method is widely used, it has two drawbacks. First, it cannot ensure that the predicted trajectories serve as a solution to the embedded QP model of DMPC, e.g., whether they satisfy the constraints or can minimize costs. This may make the predicted trajectories not physically meaningful, leading to estimation errors that can be detrimental to the control performance. 
Second, this classical method ignores the useful structural information of the DMPC model. Since UAVs use the same DMPC model, they know perfectly the dynamics and decision processes of other UAVs. Specifically, if UAV $i$ knows all the input of UAV $j$ to the DMPC model, UAV $i$ can accurately obtain the planned trajectories of UAV $j$ simply by solving the DMPC model with UAV $j$'s input. 

We will address these two drawbacks by employing a KKT-informed training approach that brings information about the DMPC to the training process. The structure of the training process is illustrated in Figure~\ref{fig:method_KKT_training}. 
We first instantiate the embedded QP in the DMPC-based controller for UAV $j$, whereby the input is the predicted trajectories of all neighbors of UAV $j$.

With the instantiated QP of UAV $j$, UAV $i$ can solve the QP of UAV $j$ and obtain a primal solution $\tilde{\bm{U}}_{*, j}^{k}$ and a dual solution $\tilde{\bm{\lambda}}_{*, j}^{k}$. Ideally, the primal solution $\tilde{\bm{U}}_{*, j}^{k}$ should be close to the predicted trajectory $\tilde{\bm{U}}_{j}^{k}$ calculated by the trajectory predictor. We leverage this insight via a KKT-informed training approach, which uses the primal solution $\tilde{\bm{U}}_{*, j}^{k}$ and dual solution $\tilde{\bm{\lambda}}_{*, j}^{k}$ to inform the training of the network parameters. Specifically, we aim to minimize the loss function written as 
\begin{align}
    \mathcal{L}_{\text{KKT}}(\tilde{\bm{U}}_{j}^{k}, \bm{U}_{j}^{k}) & = \alpha||\tilde{\bm{U}}_{j}^{k}-\bm{U}_{j}^{k}||_2^2  + \beta ||\tilde{\bm{U}}_{j}^{k}-\tilde{\bm{U}}_{*, j}^{k}||_2^2  \label{eq:kkt_loss}
\end{align}
where the first term aims to minimize the mean square error between the true (i.e., $\bm{U}_{j}^{k}$) and estimated values of the planned trajectories produced the EG-based trajectory predictor (i.e., $\tilde{\bm{U}}_{j}^{k}$), and the second term aims to minimize the mean square error between the QP solution (i.e., $\tilde{\bm{U}}_{*, j}^{k}$) and the estimated planned trajectory (i.e., $\tilde{\bm{U}}_{j}^{k}$).

When calculating the gradient of the loss function with respect to the neural network parameters, we need to backpropagate the gradient information of the primal and dual solutions. Specifically, these derivatives are found using Proposition~\ref{prp:KKT}, which performs sensitivity analysis on the KKT conditions. 

\begin{proposition}[Differentiable KKT Layer \cite{amos2017optnet}] \label{prp:KKT}
Given primal solution $\bm{w}^*$ and dual solution $\bm{\lambda}^*$ to the following QP
\begin{subequations}
    \begin{align}
        \min_{\bm{w}}\quad & \frac{1}{2}\bm{w}^T\bm{Q}(\bm{u})\bm{w} + \bm{q}(\bm{u})^T\bm{w} \\
        & \bm{G}(\bm{u})\bm{w} \leq \bm{h}(\bm{u}) \\
        & \bm{R}(\bm{u})\bm{w} = \bm{b}(\bm{u})  
    \end{align}
\end{subequations}
    where the parameters are functions with respect to parameters $\bm{u}$. Then the derivatives of the primal and dual solutions $(\frac{\partial \bm{w}^*}{\partial \bm{u}^*}, \frac{\partial \bm{\lambda}^*}{\partial \bm{u}^*})$ satisfy 
\begin{subequations}
\begin{align}
    &\bm{Q}\frac{\partial \bm{w}^*}{\partial \bm{u}^*} + \frac{\partial \bm{Q}}{\partial \bm{u}^*}\bm{w}^*+\frac{\partial \bm{q}}{\partial \bm{u}^*} \nonumber \\ & +\bm{R}\frac{\partial \bm{w}^*}{\partial \bm{u}^*}+\frac{\partial \bm{R}}{\partial \bm{u}^*}\bm{w}^*+ \bm{G}\frac{\partial {\lambda}^*}{\partial \bm{u}^*}+ \frac{\partial \bm{G}}{\partial \bm{u}^*}{\lambda}^*  = {0} \label{eq:KKT1}\\
    & \frac{\partial \bm{R}}{\partial \bm{u}^*}\bm{w}^* 
    + \bm{R}\frac{\partial \bm{w}^*}{\partial \bm{u}^*} -\frac{\partial \bm{b}^*}{\partial \bm{u}^*} = 0 \label{eq:KKT2} \\
    &\left(\frac{\partial {\lambda}^*}{\partial \bm{u}^*}\right)^T(\bm{G}\bm{w}^*-\bm{h})  \notag \\
    & + \left({\lambda}^*\right)^T\left(\frac{\partial \bm{G}}{\partial \bm{u}^*}\bm{w}^* 
    + \bm{G}\frac{\partial \bm{w}^*}{\partial \bm{u}^*} -\frac{\partial \bm{h}^*}{\partial \bm{u}^*}\right) = 0  \label{eq:KKT3}
\end{align}
\end{subequations}
\end{proposition}

Notice that the KKT layer is used solely to facilitate the training of the trajectory prediction network rather than as an additional estimator. This is because the forward propagation of the KKT layer requires solving a sequence of QPs, which can be computationally expensive for real-time implementation. 
In this way, the proposed KKT-informed training process can be applied to different trajectory predictor settings (e.g., with or without VAE-based compression and reconstruction module). 
Nevertheless, minimizing the loss function $ \mathcal{L}_{\text{KKT}}(\cdot)$ can incorporate the structural information of the QP model into the trained neural network. Hence, this treatment is expected to improve the accuracy of the prediction.

\section{Simulation and Analysis} \label{sec:simulation}
In this section, we perform simulations to evaluate the proposed trajectory prediction algorithm in a typical funnel-like environment. We first introduce the benchmark algorithms utilized in the simulations in Section~\ref{sec:benchmark}. Then the scenario generation process and parameters are given in Section~\ref{sec:scenario}. We next analyze the simulation results, including the comparison of the proposed method with other benchmark algorithms (Section~\ref{sec:performance}), the robustness in scenarios with limited communication capabilities (Section~\ref{sec:robustness}), and the sensitivity to measurement noises (Section~\ref{sec:noise}). 

\subsection{Benchmarks} 
\label{sec:benchmark}
We evaluate our proposed trajectory prediction algorithm by comparing it to the following benchmarks. These nine benchmark algorithms are chosen to evaluate the value of each component of the proposed algorithm. 
\begin{itemize}
    \item LIA.
    This benchmark follows \cite{zhu2021learning}, which is a learning interaction-aware (LIA) trajectory prediction algorithm dedicatedly designed for UAV swarm trajectory prediction task. The input to the algorithm is the real-time and historical location information of all UAVs in the swarm, as well as geometric information about the static obstacles. Notice that this algorithm does not consider the dynamic nature of the UAV alignment. The architecture of the LIA algorithm includes an LSTM-based encoder and an LSTM-based decoder.
    \item Y-Net.
    This benchmark follows \cite{mangalam2021goals}, a state-of-the-art trajectory prediction algorithm that handles the uncertainty of future trajectories with factorized goal and path multimodalities, which is widely recognized in various trajectory prediction tasks~\cite{sharma2022pedestrian,teeti2022vision}. We extend this work from its original 2D domain ($x$ and $y$) to our 3D domain ($x$, $y$, and $z$). The input to the algorithm consists of real-time and historical location information of predicted UAVs in the swarm, as well as map information about static obstacles. UAV dynamics are not considered in this benchmark. The architecture of this benchmark includes a Convolutional Neural Network (CNN)-based encoder and two CNN-based decoders.
    \item EG only. This benchmark only involves the EvolveGCN-based trajectory prediction component that uses real-time and historical location information about the surrounding UAVs to estimate their planned trajectories. This benchmark corresponds to the scenario where communication is not available. This benchmark is trained by minimizing the $l_2$ loss. The comparison between EG and LIA/Y-Net shows the value of considering the topological information about the UAV swarm. 
    \item VAE only. This benchmark only has the VAE component, where each UAV uses the decoder to reconstruct the compressed messages without integrating the real-time information about surrounding UAVs. This benchmark is trained by minimizing the $l_2$ loss. 
    \item VAE + EG (Bayesian). This benchmark assumes the availability of both VAE and the EvolveGCN-based trajectory prediction components, and uses the Bayesian formula to fuse the estimation outcomes. This benchmark is trained by minimizing the $l_2$ loss. The comparison between VAE and VAE+EG (Bayesian) shows the value of real-time sensing, and the comparison between EG and VAE+EG (Bayesian) shows the value of communication. 
    \item EG + KKT. This benchmark evaluates the estimation performance of the EG component with the aid of the KKT-informed component, where the KKT-informed component is only activated during the training process due to its forward computational complexity. The comparison between EG + KKT and EG shows the value of the differentiable KKT layer in the training of the model.
    \item VAE + KKT. This benchmark evaluates the estimation performance of the VAE component with the aid of the KKT-informed component, and the KKT-informed component is only activated during the training process. 
    \item VAE + EG + KKT. The only difference between this benchmark and VAE + EG (Bayesian) is that this benchmark is trained with the differential KKT layer. This benchmark is further evaluated under scenarios with various compression levels, communication frequencies, and packet loss rates. The comparison between VAE + EG + KKT and VAE+EG (Bayesian) shows the value of the differentiable KKT layer in the training of the model. 
    \item DMPC. This benchmark assumes perfect communication between agents and hence is treated as the oracle. The resulting planned trajectories are regarded as ground truths. 
\end{itemize}

\subsection{Implementation details}
\label{sec:scenario}
We randomly generate simulation scenarios for training and testing the proposed trajectory prediction algorithm. Specifically, each scenario simulates the dynamics of a UAV swarm in a 3-dimensional funnel-like environment with two obstacles modeled as ellipsoids. The centers of these ellipsoids are sampled uniformly at random from cubes $\mathcal{C}_1 = \{(p_x,p_y,p_z)|p_x\in[10,14], p_y\in [3.5,4.5], p_z \in [-2,2]\}$ and $\mathcal{C}_2 = \{(p_x,p_y,p_z)|p_x\in[10,14], p_y\in [-3.5,-4.5], p_z \in [-2,2]\}$, where the values are in meters. The lengths of the three axes of the ellipsoids are sampled uniformly at random from $\left\{(r_x,r_y,r_z)|r_x\in[4,8], r_y\in [3,4], r_z \in [4,8]\right\}$ (in meters). 
We further enforce the minimum distance between these two ellipsoids to range between 0.6\,m and 1.2\,m by removing the scenarios that violate this condition. 

The swarm size is sampled uniformly at random from set $N = \{n \in \mathbb{Z} | 6 \leq n \leq 14\}$. The initial locations of the UAVs are generated within a space $\{(p_x,p_y,p_z)|p_x\in[0,3], p_y\in [-4,4], p_z \in [-1,1]\}$ (in meters) using the Poisson Disk Sampling approach, which ensures that the distance between any pair of UAVs is above a minimum Euclidean distance of 0.7\,m. The minimum, maximum, and desired velocities of each UAV are assumed to be -1.5\,m/s, 1.5\,m/s, and 0.5\,m/s, respectively. 
The minimum and maximum acceleration rates are assumed to be -1\,m/s$^{2}$ and 1\,m/s$^{2}$, respectively. 
We aim to prevent collisions between UAV~$i$ and other UAVs/obstacles by requiring $||\bm{p}_i-\bm{p}||_{\bm{E}}\geq r_{\min}$, where $\bm{E}$ represents a linear transformation matrix, $\bm{p}_i$ represents the location of UAV $i$,  $\bm{p}$ represents the location of another UAV or the closest point of an obstacle, and $r_{\min}$ represents a safety buffer chosen as $r_{\min}=0.15$\,m. If $||\bm{p}_i(k)-\bm{p}(k)||_{\bm{E}}<r_{\text{coll}}$, we say that collisions occur, where the collision threshold $r_{\text{coll}}$ is chosen as 0.07\,m.  

The dynamics of the UAV swarm follow Eq.(\ref{eq:dynamics}) with the control input determined using the DMPC module, where the system matrices of the UAVs (i.e., $\bm{A}_i$ and $\bm{B}_i$) follow the dynamics of the Crazyflie 2.1 quadrotor with an underlying position controller.  The sampling time and DMPC planning horizon are $\Delta T=0.2$\,s and $P=16$, respectively. The UAV trajectories within each planning horizon are parameterized as $l=3$ Bézier curves with order $d=5$. 
The weights of objective function Eq. (\ref{eq:obj}), $q_\text{mig}, l_\text{saf}, q_\text{saf}, l_\text{coh}, q_\text{coh}, q_\text{eft}$, are set to be $1, 1e4, 1e2, 1e3, 10, 70$ respectively.
For each scenario, we perform a simulation for $T=30$\,s. Following~\cite{soria2021distributed}, we consider Gaussian position measurement noises with zero mean and a default standard deviation of 0.004\,m, which will be increased in the testing phase to evaluate the algorithm's robustness. 

We generate the training dataset by randomly generating 300 simulation scenarios according to the aforementioned distributions. Within each scenario, we collect the following information: (1) real-time locations and velocities of each UAV, (2) the location of obstacles, and (3) the QP matrices and solutions of each UAV at each time step. In total, we collected 450,000 samples, which were then transformed into the training data of the neural networks. Note that the training dataset is generated with the benchmark DMPC, meaning that we assume perfect communication channels such that the input to the QP models is always available and accurate. We only collected data from episodes in which the swarm successfully reached the goal, as this implies the successful achievement of the control aim.

The experiments are conducted on a computer with an Intel i7-12700 CPU@2.10 GHz and an NVIDIA RTX 3070 GPU. The designed learning networks are implemented in Python using Pytorch 1.13.1. The DMPC control scheme is implemented on MATLAB R2022b and the UAV swarm is simulated using Swarmlab~\cite{soria2020swarmlab}. The input size of the VAE module is 16, consistent with the prediction horizon $P$. The input size for all EG modules and LSTM encoder modules is 20, which is the horizon length for UAV past state observation $H$. All hidden layers in the neural network have 128 neurons and the output layer of different modules are all set to be 16. ReLu is selected to be the activate function and all the neural networks are trained under a learning rate of 0.0001. 

\subsection{Performance in testing scenarios}
\label{sec:performance}
In this subsection, we show the results of the testing scenarios. Specifically, we generate another set of 50 scenarios for testing, according to the scenario distribution described in Section~\ref{sec:scenario}. For each testing scenario, we are interested in analyzing control performance and prediction performance. To evaluate control performance, we measure two key factors for each benchmark algorithm: the total incurred cost throughout the simulation period for efficiency and the distances between agents as well as between agents and obstacles for safety.
We will further evaluate our algorithm in various scenarios with different parameters to show the robustness of our algorithms in less ideal cases. The data compression levels of VAE, i.e., the latent dimension, range from $L=6$ to $L=42$. Notice that the original dimension of the planned trajectories is $L=48$. The communication frequencies range from $f_{\text{comm}}=5$\,Hz to $f_{\text{comm}}=1.25$\,Hz. The packet loss rates are set to range from $P_{\text{loss}}=0$ to $P_{\text{loss}}=1/2$, and the measurement noise levels range from $0$ to $0.032$. 
For prediction performance, we will evaluate the mean square errors (MSE) of the predicted planned trajectories compared to the ground-truth planned trajectories solved by DMPC. 
During the testing phase, the UAV swarm consistently reaches its goal under different benchmarks.

\begin{table}[htbp] 
\caption{Cost comparison between different algorithms, with costs in units of $10^5$}
\begin{tabular}{p{1.0cm}|p{0.8cm}|p{0.8cm}|p{0.8cm}|p{0.8cm}|p{0.8cm}|p{0.8cm}}
\hline\hline
Algorithm   & Safe-Agent Cost   & Safe-Obs Cost   & Migra-tion Cost   & Control Effort   & Cohesion Cost   & Total Cost\\\hline 
LIA      & 0.0544      & 0.638      & 1.09      & 1.17      & 0.450      & 3.40      \\\hline
Y-Net    & 0      & 0      & 1.85      & 2.04      & 2.38      & 6.27      \\\hline
EG       & 0      & 0      & 1.19      & 1.58      & 1.87      & 4.65      \\\hline
VAE      & 0      & 0      & 1.17      & 1.42      & 1.79      & 4.39      \\\hline
EG + KKT      & 0      & 0      & 1.75      & 1.91      & 0.544      & 4.20      \\\hline
VAE + EG (Bayesian)        & 0           & 0           & 1.22      & 1.86      & 1.10      & 4.18      \\\hline
VAE + KKT         & 0      & 0      & 1.15      & 1.34      & 0.971      & 3.46      \\\hline
VAE + EG + KKT    & 0      & 0      & 1.18      & 1.51      & 0.615      & 3.30      \\\hline
DMPC      & 0     & 0      & 1.16      & 1.41      & 0.673      & 3.24      \\\hline\hline
\end{tabular}\label{tab:benchmark}
\end{table}

\begin{figure}[htbp]
    \begin{subfigure}[t]{0.48\textwidth} 
        \includegraphics[width=\textwidth]{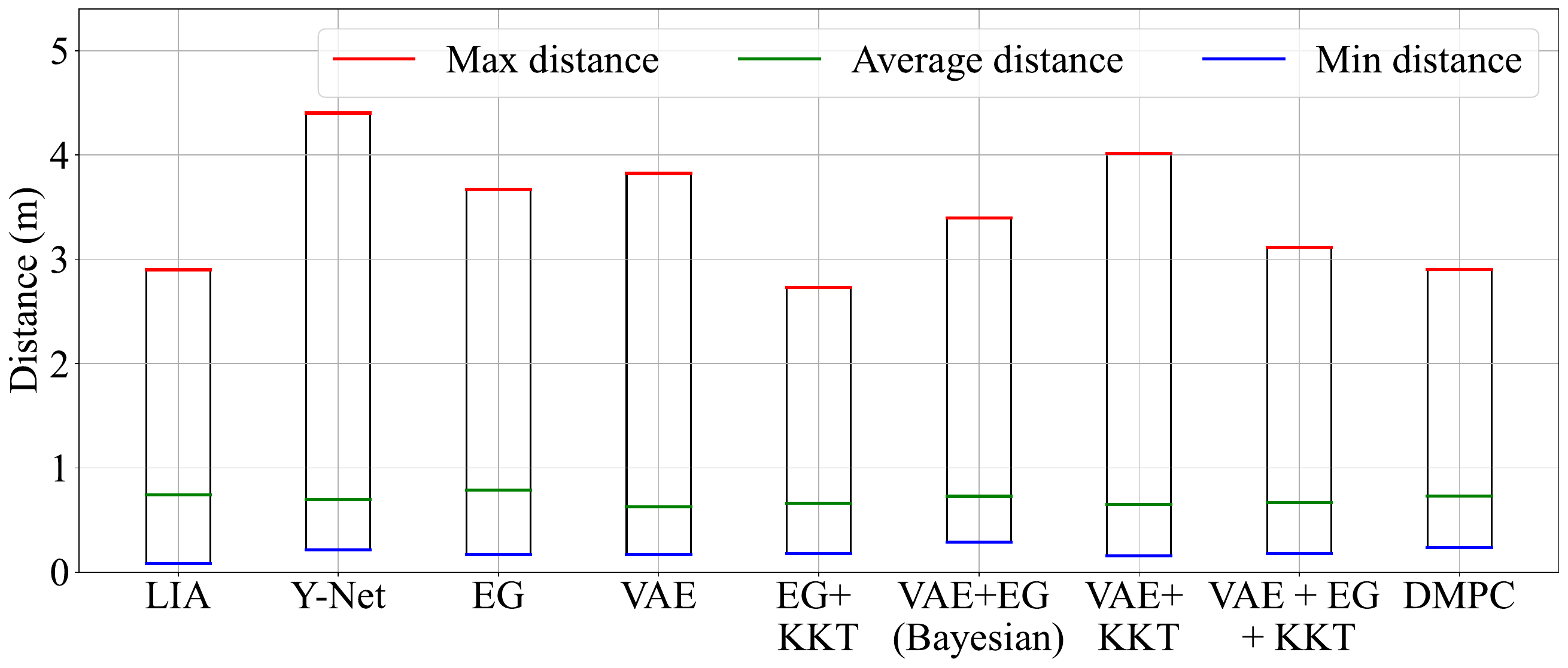}
        \caption{Agent-agent distance comparison of different benchmarks.}
        \label{fig:Comparison_agent_agent}
    \end{subfigure}\hfill
    \begin{subfigure}[t]{0.48\textwidth} 
        \includegraphics[width=\textwidth]{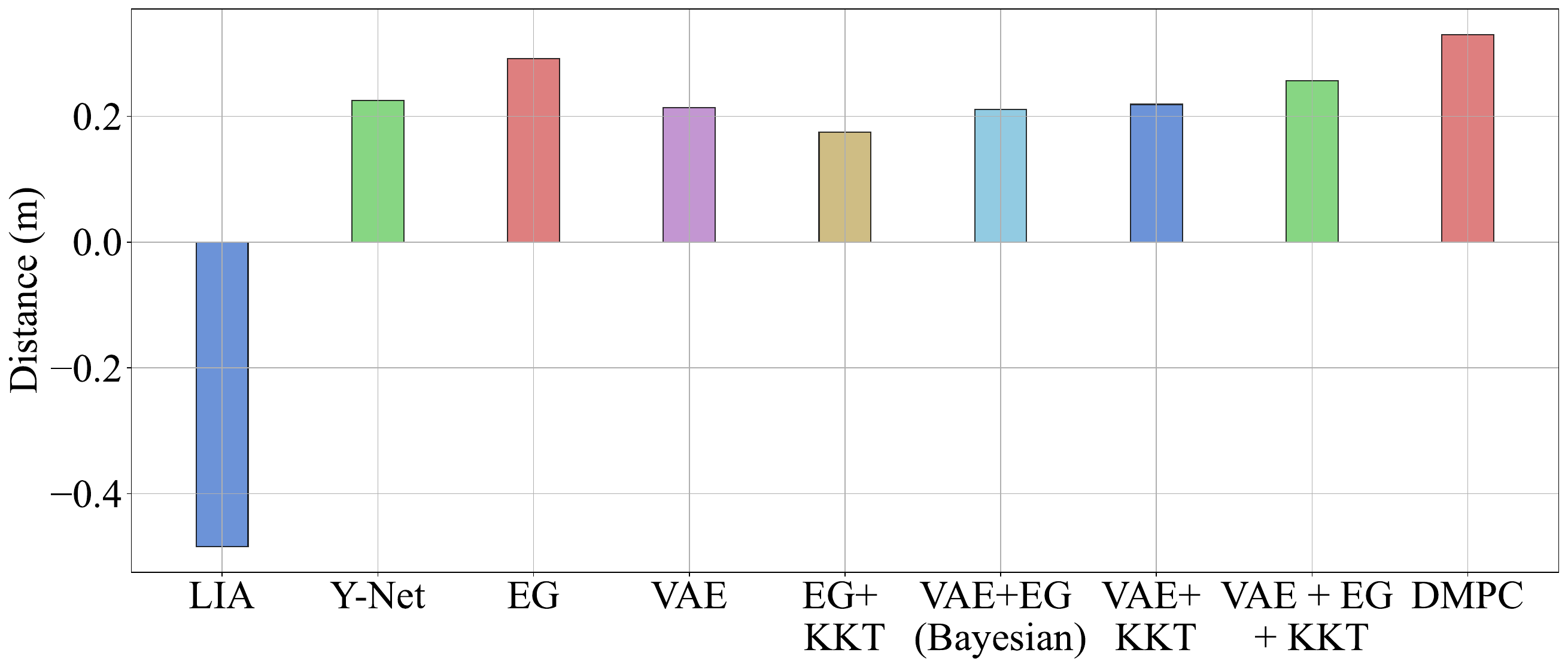}
        \caption{Minimum agent-obstacle distance of different benchmarks.}
        \label{fig:Comparison_agent_obstacle}
    \end{subfigure}
    \caption{Safe distance comparison of different benchmarks.} 
    \label{fig:Comparison_both} 
\end{figure}

\begin{figure*}[hbp]
    \centering

    \begin{subfigure}[b]{0.49\textwidth}
        \includegraphics[width=\textwidth]{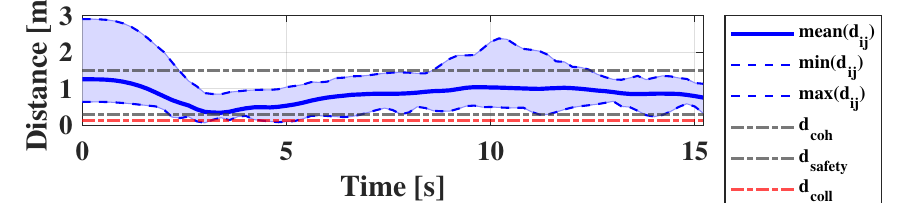}
        \caption{Distance between UAVs with LIA applied.}
    \end{subfigure}
    \begin{subfigure}[b]{0.49\textwidth}
        \includegraphics[width=\textwidth]{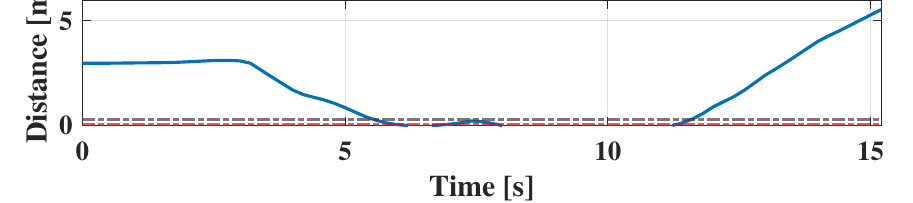}  
        \caption{Minimum UAV-obstacle distance with LIA applied.}
    \end{subfigure}    
    
    \begin{subfigure}[b]{0.49\textwidth}
        \includegraphics[width=\textwidth]{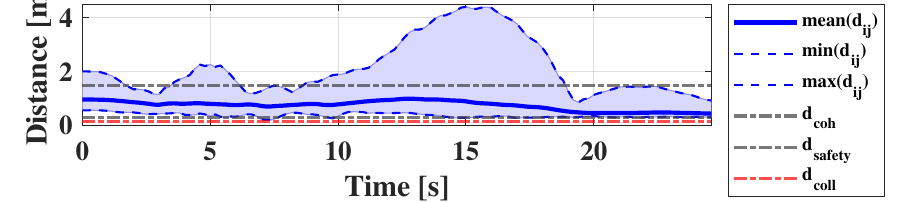}
        \caption{Distance between UAVs with Y-Net applied.}
    \end{subfigure}
    \begin{subfigure}[b]{0.49\textwidth}
        \includegraphics[width=\textwidth]{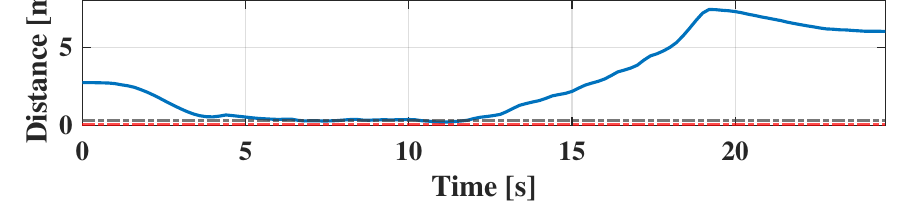}
        \caption{Minimum UAV-obstacle distance with Y-Net applied.}
    \end{subfigure}    

    \begin{subfigure}[b]{0.49\textwidth}
        \includegraphics[width=\textwidth]{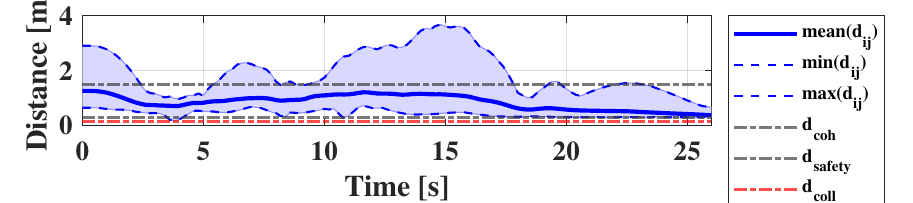}
        \caption{Distance between UAVs with EG applied.}
    \end{subfigure}
    \begin{subfigure}[b]{0.49\textwidth}
        \includegraphics[width=\textwidth]{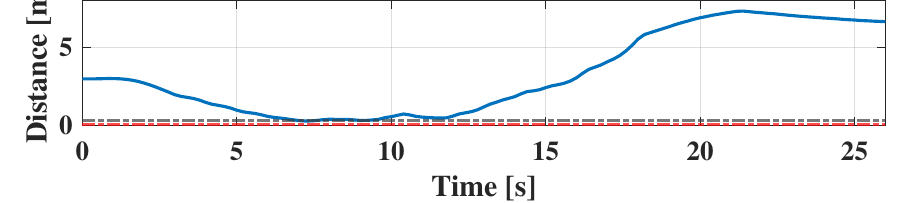}
        \caption{Minimum UAV-obstacle distance with EG applied.}
    \end{subfigure}    

    \begin{subfigure}[b]{0.49\textwidth}
        \includegraphics[width=\textwidth]{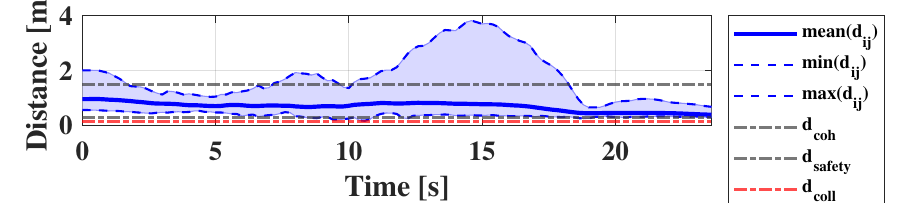}
        \caption{Distance between UAVs with VAE applied.}
    \end{subfigure}
    \begin{subfigure}[b]{0.49\textwidth}
        \includegraphics[width=\textwidth]{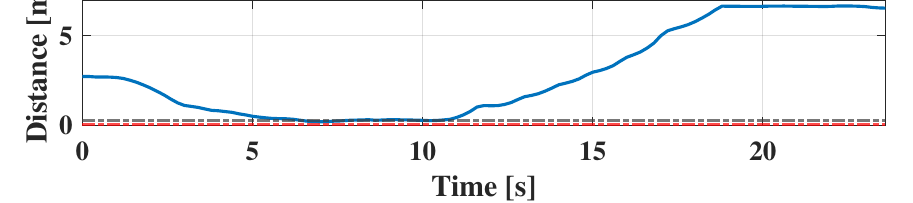}
        \caption{Minimum UAV-obstacle distance with VAE applied.}
    \end{subfigure}    

    \begin{subfigure}[b]{0.49\textwidth}
        \includegraphics[width=\textwidth]{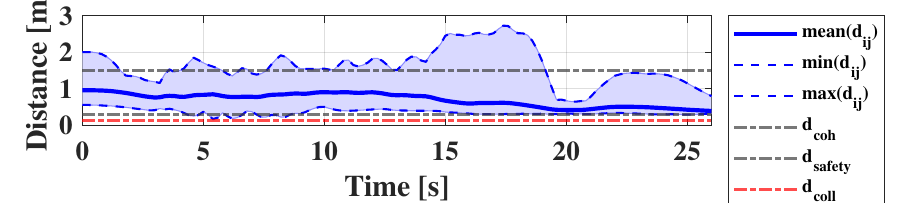}
        \caption{Distance between UAVs with EG + KKT applied.}
    \end{subfigure}
    \begin{subfigure}[b]{0.49\textwidth}
        \includegraphics[width=\textwidth]{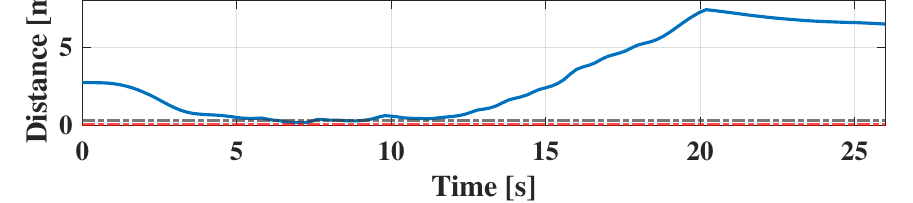}
        \caption{Minimum UAV-obstacle distance with EG + KKT applied.}
    \end{subfigure}    

    \begin{subfigure}[b]{0.49\textwidth}
        \includegraphics[width=\textwidth]{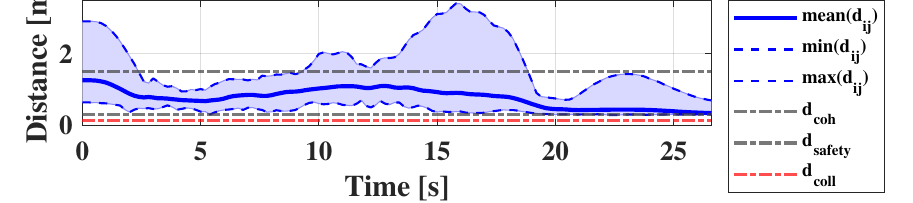}
        \caption{Distance between UAVs with VAE+EG (Bayesian) applied.}
    \end{subfigure}
    \begin{subfigure}[b]{0.49\textwidth}
        \includegraphics[width=\textwidth]{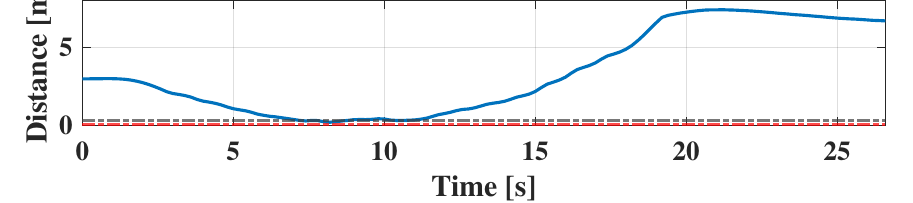}
        \caption{Minimum UAV-obstacle distance with VAE+EG (Bayesian) applied.}
    \end{subfigure}   
    
    \begin{subfigure}[b]{0.49\textwidth}
        \includegraphics[width=\textwidth]{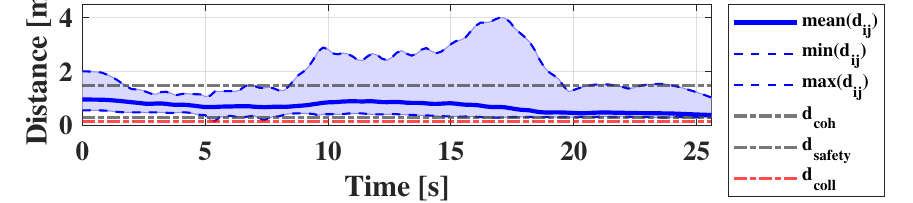}
        \caption{Distance between UAVs with VAE + KKT applied.}
    \end{subfigure}
    \begin{subfigure}[b]{0.49\textwidth}
        \includegraphics[width=\textwidth]{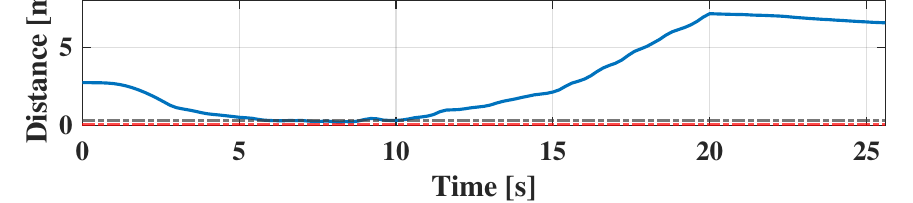}
        \caption{Minimum UAV-obstacle distance with VAE + KKT applied.}
    \end{subfigure}  
    
    \begin{subfigure}[b]{0.49\textwidth}
        \includegraphics[width=\textwidth]{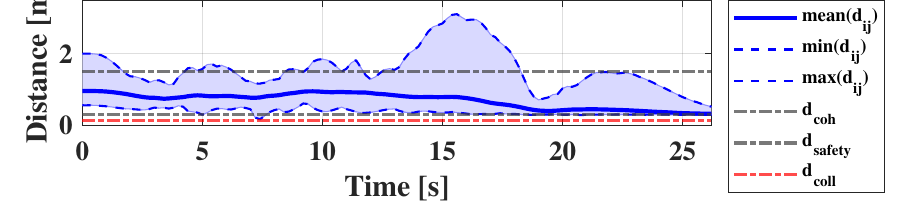}
        \caption{Distance between UAVs with VAE+EG+KKT applied.}
    \end{subfigure}
    \begin{subfigure}[b]{0.49\textwidth}
        \includegraphics[width=\textwidth]{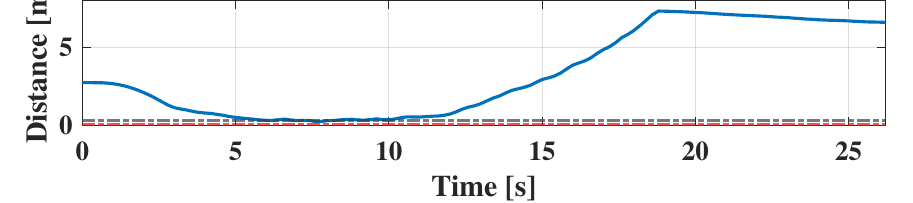}
        \caption{Minimum UAV-obstacle distance with VAE+EG+KKT applied.}
    \end{subfigure}    

        \begin{subfigure}[b]{0.49\textwidth}
        \includegraphics[width=\textwidth]{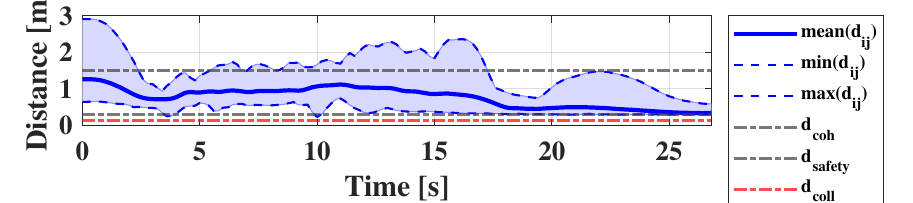}
        \caption{Distance between UAVs with DMPC applied.}
    \end{subfigure}
    \begin{subfigure}[b]{0.49\textwidth}
        \includegraphics[width=\textwidth]{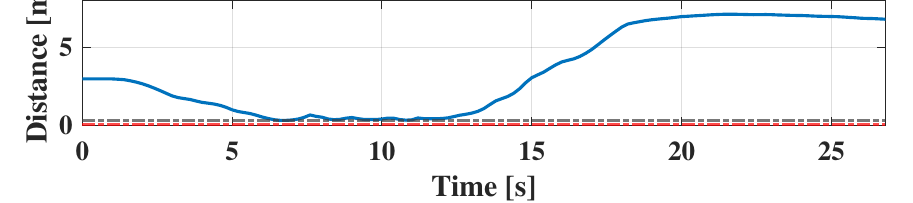}
        \caption{Minimum UAV-obstacle distance with DMPC applied.}
    \end{subfigure}

    \caption{Evolution of distances with different benchmarks applied.}
    \label{fig:distance_benchmarks}
\end{figure*}

We first compare the performance of the benchmarks listed in Section~\ref{sec:benchmark} with all parameters set to the default values, i.e.,  $L=24$, $f_{\text{comm}}=5$\,Hz, $P_{\text{loss}}=0$, and measurement noise level 0.004\,m. The resulting average control costs across all scenarios are summarized in Table~\ref{tab:benchmark}, whereby the Safe-Agent Cost penalizes the violation of the safety constraints between UAVs, the Safe-Agent Obs Cost penalizes the violation of the safety constraints between UAVs and obstacles, the Migration Cost penalizes the distance to the pre-defined migration point, the Control Effort cost helps to minimize the energy spent to execute the control action, the Cohesion Cost penalizes UAVs being too far away from each other, and the total cost is the sum of all five costs.  
Figure~\ref{fig:Comparison_both} illustrate agent-agent and agent-obstacle distances, averaged over all testing scenarios, whereas the temporal evolution of these distances in a typical scenario is illustrated in Figure~\ref{fig:distance_benchmarks}. 
Table~\ref{tab:prediction accuracy} summarizes the average trajectory prediction errors (measured as MSE along the $x$, $y$, and $z$ axes using the Euclidean norm) across all testing scenarios.

\emph{Performance of the proposed algorithm}. From Table \ref{tab:benchmark} and Figure \ref{fig:Comparison_both}, we can see that the proposed trajectory prediction algorithm VAE+EG+KKT ensures safety and yields similar control performance to DMPC, outperforming other benchmarks. This can be further elaborated by looking at the prediction errors resulting from VAE+EG+KKT, which is close to 0 and outperforms other algorithms. This shows that the proposed trajectory prediction algorithm is effective and can lead to a close-to-optimal performance, even when reducing the dimension of the communicated message by half.

We next demonstrate the value of each component in the proposed algorithm by comparing groups of benchmark algorithms. 

\emph{Value of the EvolveGCN component in considering time-varying UAV alignment}. We demonstrate the value of incorporating the EvolveGCN component by comparing LIA, Y-Net and EG. 
From Table~\ref{tab:benchmark}, Figure \ref{fig:Comparison_both}, Figure~\ref{fig:distance_benchmarks}a) -- Figure~\ref{fig:distance_benchmarks}f), 
we can see that although LIA outperforms Y-Net and EG in terms of total costs, it cannot ensure safety. It is shown in Figure~\ref{fig:distance_benchmarks}a) and ~\ref{fig:distance_benchmarks}b) that there are collisions both between UAVs and between UAVs and the obstacles. Moreover, we find that Y-Net maintains the largest cost in the migration cost, control effort and cohesion cost, consequently resulting in the largest total cost. It is shown in Figure~\ref{fig:distance_benchmarks}c) and~\ref{fig:distance_benchmarks}d) that the maximum distance between agents is high, while the minimum distance between agents and obstacles is rather low. The reason can be attributed to the fact that LIA and Y-Net do not produce accurate trajectory prediction outcomes, as can be seen in Table~\ref{tab:prediction accuracy} where the prediction MSE is higher than other benchmarks. This indicates that LIA and Y-Net may not perform well in dynamic environments where the layout of the UAV swarm constantly changes. This also highlights the benefits of the EvolveGCN framework in improving prediction accuracy.

\emph{Value of the KKT-informed training procedure}. We evaluate the benefits of the KKT-informed training process by comparing VAE+EG+KKT with VAE+EG (Bayesian). We can see that although the prediction performance between the two benchmarks is the same (as shown in Tables~\ref{tab:prediction accuracy}), the control performance of VAE+EG+KKT is significantly better than VAE+EG (Bayesian) (as shown in Tables~\ref{tab:benchmark},Figure \ref{fig:Comparison_both}, Figure~\ref{fig:distance_benchmarks}k)-~\ref{fig:distance_benchmarks})p). This is because the KKT-informed training process enables the prediction outcome of VAE+EG+KKT to better align with the embedded QP in DMPC. 
This can also be reflected by comparing EG with EG+KKT and VAE with VAE+KKT, which use the KKT-informed component to enhance the performance of both EG and VAE. The only difference is that EG+KKT and VAE+KKT integrate the KKT-informed layer during the training procedure.
We can see from Tables~\ref{tab:benchmark}, Figure \ref{fig:Comparison_both}, Figure~\ref{fig:distance_benchmarks}e)-~\ref{fig:distance_benchmarks}n) that using KKT alone can enhance the performance of individual predictors. Notice that due to its high computational burden in forward propagation, we recommend using the KKT component only in the training procedure.

\emph{Value of communication}. We evaluate the benefits of having the communicated message by comparing VAE+EG (Bayesian) and EG, which shows the value of having the reconstructed trajectories from VAE. Our conclusion is similar to the above analysis in that VAE+EG (Bayesian) significantly outperforms EG, as demonstrated in Tables~\ref{tab:benchmark}--\ref{tab:prediction accuracy}, as well as in Figure \ref{fig:Comparison_both} and Figures~\ref{fig:distance_benchmarks}e)-\ref{fig:distance_benchmarks}l). 

\begin{table}[htbp] 
\caption{Trajectory prediction accuracy (meter)}
\centering
\begin{tabular}{p{0.5cm}|p{0.575cm}|p{0.575cm}|p{0.575cm}|p{0.575cm}|p{0.575cm}|p{0.575cm}|p{0.575cm}|p{0.575cm}}
\hline\hline
Time Step & LIA  & Y-Net  & EG  & VAE  & EG + KKT  & VAE + EG (Baye-sian)  & VAE + KKT  & VAE + EG + KKT \\
1  & 0.49 & 0.37     & 0.34 & 0.18   & 0.14      & 0.07    & 0.06    & 0.06        \\\hline
2  & 0.50 & 0.35     & 0.30 & 0.15   & 0.12      & 0.05    & 0.07    & 0.06        \\\hline
3  & 0.59 & 0.37     & 0.28 & 0.14   & 0.12      & 0.05    & 0.05    & 0.05        \\\hline
4  & 0.65 & 0.33     & 0.27 & 0.14   & 0.12      & 0.05    & 0.05    & 0.05        \\\hline
5  & 0.74 & 0.36     & 0.26 & 0.14   & 0.11      & 0.05    & 0.05    & 0.04        \\\hline
6  & 0.81 & 0.37     & 0.28 & 0.15   & 0.13      & 0.05    & 0.05    & 0.05        \\\hline
7  & 0.88 & 0.42     & 0.28 & 0.17   & 0.15      & 0.06    & 0.07    & 0.06        \\\hline
8  & 0.98 & 0.44     & 0.30 & 0.18   & 0.16      & 0.07    & 0.08    & 0.07        \\\hline
9  & 1.05 & 0.45     & 0.32 & 0.19   & 0.17      & 0.08    & 0.08    & 0.08        \\\hline
10 & 1.13 & 0.46     & 0.33 & 0.20   & 0.17      & 0.09    & 0.08    & 0.10        \\\hline
11 & 1.19 & 0.40     & 0.35 & 0.22   & 0.19      & 0.10    & 0.09    & 0.10        \\\hline
12 & 1.27 & 0.44     & 0.37 & 0.24   & 0.22      & 0.10    & 0.10    & 0.11        \\\hline
13 & 1.35 & 0.40     & 0.38 & 0.25   & 0.22      & 0.10    & 0.11    & 0.10        \\\hline
14 & 1.42 & 0.43     & 0.39 & 0.27   & 0.24      & 0.11    & 0.11    & 0.10        \\\hline
15 & 1.50 & 0.44     & 0.40 & 0.28   & 0.25      & 0.11    & 0.10    & 0.10        \\\hline
16 & 1.58 & 0.50     & 0.42 & 0.30   & 0.27      & 0.12    & 0.12    & 0.11        \\\hline\hline
\end{tabular} \label{tab:prediction accuracy}
\end{table}

\subsection{Robustness in scenarios with limited communication capabilities} \label{sec:robustness}

We next demonstrate the robustness of the proposed trajectory prediction algorithm in scenarios with limited communication capabilities by evaluating the sensitivity of the proposed algorithm in these scenarios. Specifically, we evaluate the impact of trajectory compression (due to limited bandwidth, energy requirements, etc.), lower communication frequency, and packet loss on the control performance. 

\emph{Impact of trajectory compression}. We evaluate our trajectory prediction algorithm with various levels of trajectory compression, represented by the dimension of the VAE latent variable $\bm{z}_k$ (i.e., the messages exchanged between UAVs). Other parameters remain at the nominal values. Table~\ref{tab:compression} shows the resulting control performance of all scenarios. Table~\ref{tab:compression_safety} shows the safety performance, where DMPC is added as a reference. As we can see from Table~\ref{tab:compression}, the control performance is close to that of DMPC as long as the latent dimension is greater than 24. For lower dimensions, although the control performance drops, the algorithm can still yield safe actions and outperform other benchmarks such as EG, VAE, and VAE+EG (Bayesian) as demonstrated in Table~\ref{tab:benchmark}. This shows that our proposed algorithm can work effectively even with low latent dimensions, suggesting that the algorithm can accommodate relatively low bandwidth.

\begin{table}[htbp] 
\caption{Sensitivity to the level of trajectory compression (control performance), with costs in units of $10^5$}
\centering
\begin{tabular}{p{1.0cm}|p{0.8cm}|p{0.8cm}|p{0.8cm}|p{0.8cm}|p{0.8cm}|p{0.8cm}}
\hline\hline
Latent Dim.  & Safe-Agent Cost  & Safe-Obs Cost  & Migra-tion Cost  & Control Effort  & Cohesion Cost  & Total Cost\\\hline 
6  & 0       & 0       & 1.21       & 1.47         & 1.30        & 3.99     \\\hline
18 & 0       & 0       & 1.19       & 1.06         & 1.25        & 3.49     \\\hline
24 & 0       & 0       & 1.18       & 1.51         & 0.615       & 3.30     \\\hline
30 & 0       & 0       & 1.24       & 1.51         & 0.533       & 3.27     \\\hline
DMPC  (48)   & 0       & 0          & 1.16         & 1.41        & 0.673    & 3.24 \\\hline\hline
\end{tabular}\label{tab:compression}
\end{table}

\begin{table}[htbp] 
\caption{Sensitivity to the level of trajectory compression  (safety performance)}
\centering
\begin{tabular}{p{1.5cm}|p{1.25cm}|p{1.25cm}|p{1.25cm}|p{1.25cm}}
\hline\hline
Latent Dim. & Min Agent-agent Distance & Average Agent-agent Distance & Max Agent-agent Distance & Min Agent-obstacle Distance \\\hline
6  & 0.16               & 0.70             & 3.99         & 0.18              \\\hline
18 & 0.19               & 0.65             & 3.84         & 0.28              \\\hline
24 & 0.18               & 0.66             & 3.11         & 0.26              \\\hline
30 & 0.23               & 0.70             & 2.93         & 0.21              \\\hline
DMPC  (48)              & 0.24             & 0.73         & 2.91          & 0.33          \\\hline\hline
\end{tabular}~\label{tab:compression_safety}
\end{table}

\emph{Impact of lower communication frequencies}. We evaluate the robustness of our prediction algorithm in scenarios with various levels of communication frequencies, i.e., 1.25\,Hz, 2.5\,Hz, and 5\,Hz, corresponding to communication every 4,2,1 time steps. Other parameters remain at the nominal values. Table~\ref{tab:freq} shows the resulting control performance of all scenarios. Table~\ref{tab:freq_safety} shows the safety performance.  As we can see from Table~\ref{tab:freq}, by reducing communication frequency, the control performance drops but is still better than other benchmark algorithms such as Y-Net and EG. Moreover, we can see that the algorithm can still ensure safety, as the minimum distance between agents and between agents and obstacles is kept higher than the collision threshold. Therefore, we can moderately reduce the communication frequency while ensuring satisfactory control performance.

\begin{table}[htbp]
\caption{Sensitivity to communication frequencies (control performance), with costs in units of $10^5$}
\begin{tabular}{p{1.0cm}|p{0.8cm}|p{0.8cm}|p{0.8cm}|p{0.8cm}|p{0.8cm}|p{0.8cm}}
\hline\hline
Comm. Freq. (Hz)  & Safe-Agent Cost   & Safe-Obs Cost   & Migra-tion Cost   & Control Effort   & Cohesion Cost   & Total Cost\\\hline 
1.25    & 0    & 0       & 1.77      & 0.936      & 1.82     & 4.52    \\\hline
2.5     & 0    & 0       & 1.84      & 1.92       & 0.112     & 3.87    \\\hline
5       & 0    & 0       & 1.18      & 1.51       & 0.615     & 3.30    \\\hline\hline
\end{tabular}\label{tab:freq}
\end{table}

\begin{table}[htbp]
\caption{Sensitivity to communication frequencies (safety performance)}
\begin{tabular}{p{1.5cm}|p{1.25cm}|p{1.25cm}|p{1.25cm}|p{1.25cm}}
\hline\hline
     Comm. Freq. (Hz)           & Min Agent-agent Distance & Average Agent-agent Distance & Max Agent-agent Distance & Min Agent-obstacle Distance \\\hline
1.25                    & 0.25               & 0.73             & 2.86      & 0.30        \\\hline
2.5                     & 0.29               & 0.63             & 2.01      & 0.31        \\\hline
5                       & 0.18               & 0.66             & 3.11      & 0.26        \\\hline\hline
\end{tabular}~\label{tab:freq_safety}
\end{table}

\emph{Impact of packet loss}. We evaluate the robustness of our prediction algorithm in scenarios with various packet loss probabilities (i.e., 1/2, 1/4, 1/8, 1/16). Other parameters remain at the nominal values. 
Table~\ref{tab:packet_loss} shows the resulting control performance of all scenarios. Table~\ref{tab:packet_loss_safety} shows the safety performance. As we can see from Table~\ref{tab:packet_loss}, by increasing packet loss probabilities, the control performance drops but is still comparable with other benchmark algorithms. Note that with the same amount of communications, packet loss has a more significant impact on the control performance than reducing communication frequencies. This is because reducing communication frequencies can still lead to a well-structured communication graph that evolves according to the dynamics of UAVs, while packet loss may make the communication graph more random. Nevertheless, we can see from Table~\ref{tab:packet_loss_safety} that the algorithm can still ensure safety.

\begin{table}[htbp]
\caption{Sensitivity to packet loss probabilities (control performance), with costs in units of $10^5$}
\begin{tabular}{p{1.0cm}|p{0.8cm}|p{0.8cm}|p{0.8cm}|p{0.8cm}|p{0.8cm}|p{0.8cm}}
\hline\hline
Packet Loss Pro.   & Safe-Agent Cost   & Safe-Obs Cost   & Migra-tion Cost   & Control Effort   & Cohesion Cost   & Total Cost\\\hline 
1/2  & 0        & 0             & 1.24      & 1.39      & 1.40     & 4.02 \\\hline
1/4  & 0        & 0             & 1.28      & 1.45      & 1.10     & 3.84 \\\hline
1/8  & 0        & 0             & 1.22      & 1.39      & 1.12     & 3.73 \\\hline
1/16 & 0        & 0             & 1.26      & 1.37      & 0.784    & 3.42 \\\hline
0    & 0        & 0             & 1.18      & 1.51      & 0.615    & 3.30  \\\hline\hline
\end{tabular} \label{tab:packet_loss}
\end{table}

\begin{table}[htbp]
\caption{Sensitivity to packet loss probabilities  (safety performance)}
\begin{tabular}{p{1.5cm}|p{1.25cm}|p{1.25cm}|p{1.25cm}|p{1.25cm}}
\hline\hline
Packet loss pro.           & Min Agent-agent Distance & Average Agent-agent Distance & Max Agent-agent Distance & Min Agent-obstacle Distance \\\hline
1/2  & 0.27               & 0.68             & 3.83         & 0.33      \\\hline
1/4  & 0.23               & 0.73             & 3.73         & 0.10      \\\hline
1/8  & 0.23               & 0.67             & 3.33         & 0.29      \\\hline
1/16 & 0.19               & 0.72             & 3.46         & 0.19      \\\hline
0    & 0.18               & 0.66             & 3.11         & 0.26      \\\hline\hline
\end{tabular} \label{tab:packet_loss_safety}
\end{table}

\subsection{Sensitivity to measurement noises} \label{sec:noise}

We evaluate the robustness of our prediction algorithm in scenarios with various measurement noises ranging from 0 to 0.032\,m. Other parameters remain at the nominal values. 
Table~\ref{tab:measurement} shows the resulting control performance of all scenarios. Table~\ref{tab:measurement_safety} shows the safety performance. 
As we can see from Table~\ref{tab:measurement}, the control performance drops as the magnitude of measurement noises increases. This is expected because as the quality of communication reduces, the measurement becomes more important for UAVs to sense their surroundings. Nevertheless, we can see from Table~\ref{tab:measurement_safety} that the algorithm can still ensure safety with moderate measurement noises.

\begin{table}[htbp] 
\caption{Sensitivity to measurement noises (control performance), with costs in units of $10^5$}
\centering
\begin{tabular}{p{1.0cm}|p{0.8cm}|p{0.8cm}|p{0.8cm}|p{0.8cm}|p{0.8cm}|p{0.8cm}}
\hline\hline
Noise Level (m)  & Safe-Agent Cost   & Safe-Obs Cost   & Migra-tion Cost   & Control Effort   & Cohesion Cost   & Total Cost\\\hline
0.032 & 0      & 0             & 1.10         & 1.31         & 1.67         & 4.08 \\\hline
0.024 & 0      & 0             & 1.12         & 1.35         & 1.52         & 3.98 \\\hline
0.016 & 0      & 0             & 1.10         & 1.29         & 1.33         & 3.72 \\\hline
0.008 & 0      & 0             & 1.11         & 1.35         & 1.16         & 3.61 \\\hline
0.004 & 0      & 0             & 1.18         & 1.51         & 0.615        & 3.30  \\\hline\hline
\end{tabular}\label{tab:measurement}
\end{table}

\begin{table}[htbp] 
\caption{Sensitivity to measurement noises  (safety performance)}
\centering
\begin{tabular}{p{1.5cm}|p{1.25cm}|p{1.25cm}|p{1.25cm}|p{1.25cm}}
\hline\hline
Noise Level  (m)           & Min Agent-agent Distance & Average Agent-agent Distance & Max Agent-agent Distance & Min Agent-obstacle Distance \\\hline
0.032 & 0.14               & 0.67             & 3.56         & 0.31         \\\hline
0.024 & 0.17               & 0.74             & 3.37         & 0.12         \\\hline
0.016 & 0.23               & 0.71             & 3.62         & 0.31         \\\hline
0.008 & 0.19               & 0.74             & 3.40         & 0.22         \\\hline
0.004 & 0.18               & 0.66             & 3.11         & 0.26         \\\hline\hline
\end{tabular}\label{tab:measurement_safety}
\end{table}

\section{Conclusion} \label{sec:conclusion}
In this paper, we propose a learning-based trajectory prediction algorithm that can enable communication-efficient UAV swarm control in a cluttered environment. Specifically, our proposed algorithm can enable each UAV to predict the planned trajectories (i.e., DMPC solutions) of its neighbors in scenarios with various levels of communication capabilities. The predicted planned trajectories will serve as input to a DMPC approach with an embedded optimization problem as a QP. The proposed algorithm involves a VAE-based trajectory compression and reconstruction model and an EvolveGCN-based trajectory prediction model, which are combined via the Bayesian formula. We further develop a KKT-informed training approach that applies the KKT conditions in the training process and thus encodes DMPC information into the trained neural network. Results show that the proposed algorithm outperforms state-of-the-art benchmarks. We have further demonstrated the value of each component of the proposed algorithm and its robustness in scenarios with limited communication capabilities and measurement noises. 

This research opens several interesting directions. First, we would like to evaluate the proposed algorithm in more complex environments, e.g., with a larger swarm size and denser/less structured obstacles. Second, it would be interesting to improve the generalizability of the proposed algorithm to a broader range of possibly changing environments. One promising way of achieving this is via meta-learning or life-long learning, enabling quick adaptation and continuous updates of the proposed algorithm. Third, we would like to explicitly investigate how to encode some prior knowledge of the communication channels into the learned neural networks to better handle possible communication failures or restrictions, in order to improve the robustness of the algorithm. Fourth, it would be interesting to perform co-design of the trajectory prediction with communication configurations. 

\bibliography{ref}

\begin{IEEEbiography}
[{\includegraphics[width=1in,height=1.25in,clip,keepaspectratio]{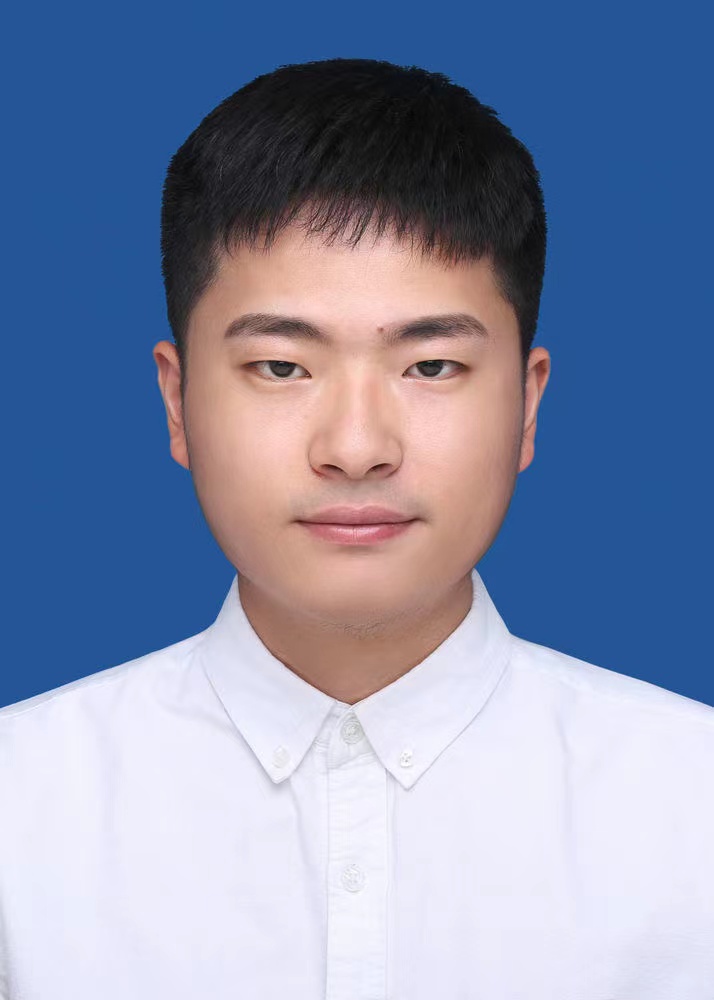}}]{Longhao Yan} receives the B.Eng. degree and M.Eng. degree in School of Electronics and Control Engineering from Chang’an University, Xi’an, China, in 2019 and 2022 respectively. He is currently working towards a Ph.D. degree with the National University of Singapore. His research interests include lateral control and trajectory prediction of intelligent transportation system.
\end{IEEEbiography}
\begin{IEEEbiography}
[{\includegraphics[width=1in,height=1.25in,clip,keepaspectratio]{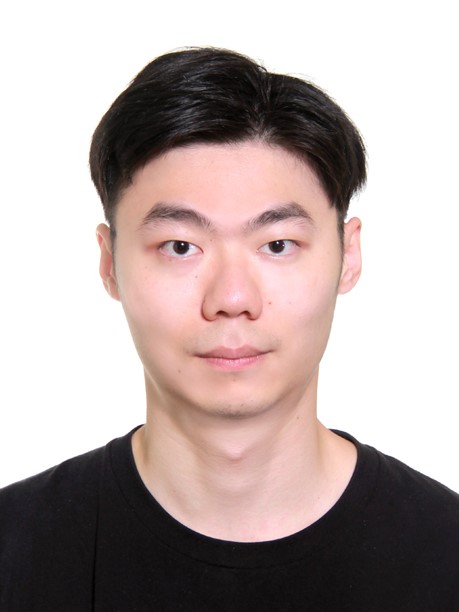}}]{Jingyuan Zhou} received the B.Eng. degree in Electronic Information Science and Technology from Sun Yat-sen University, Guangzhou, China, in 2022. He is currently working towards a Ph.D. degree with the National University of Singapore. His research interests include safe and secure connected and automated vehicles control in intelligent transportation systems.
\end{IEEEbiography}
\begin{IEEEbiography}
[{\includegraphics[width=1in,height=1.25in,clip,keepaspectratio]{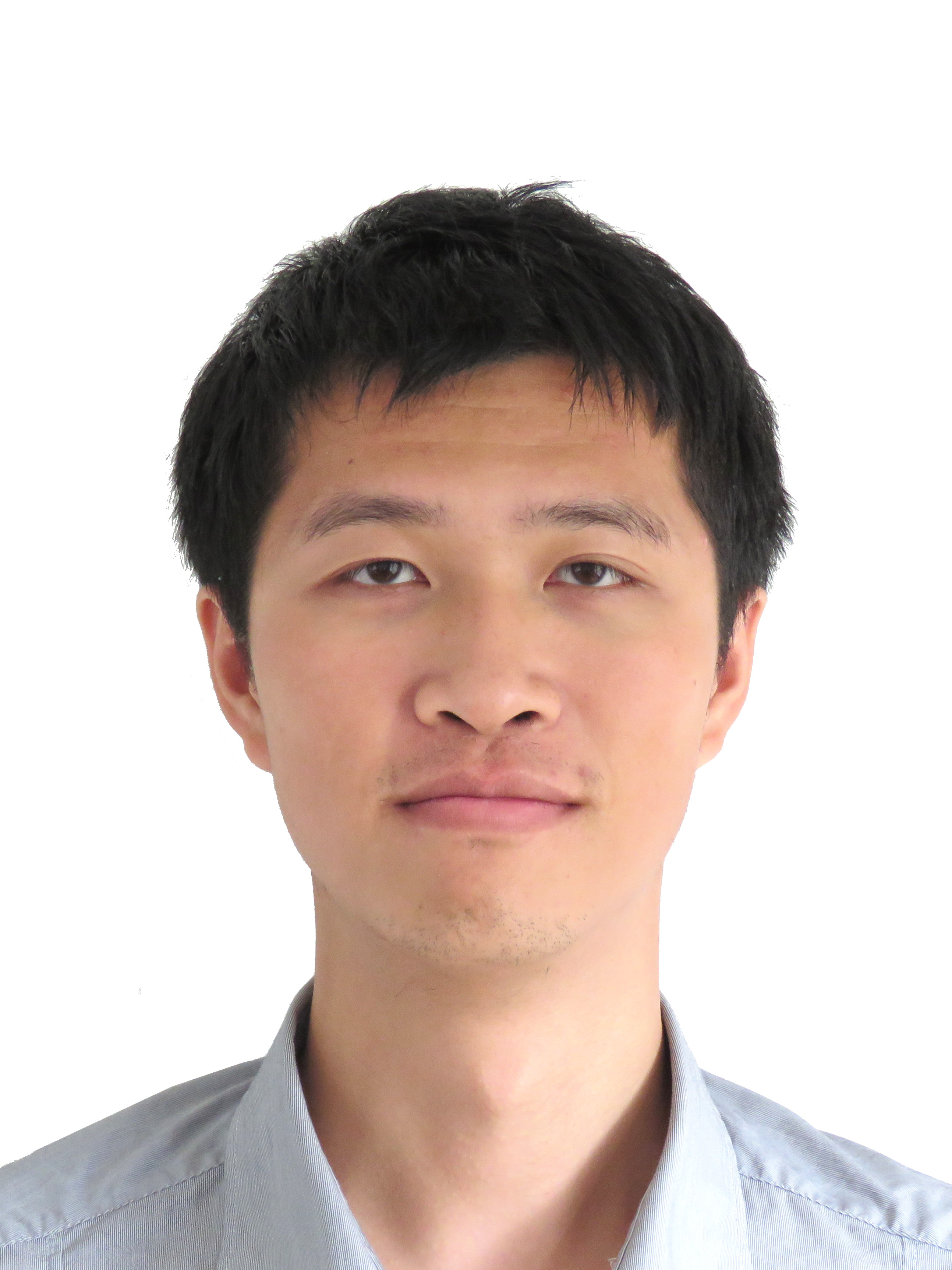}}]{Kaidi Yang}  is an Assistant Professor in the Department of Civil and Environmental Engineering at the National University of Singapore. Prior to this, he was a postdoctoral researcher with the Autonomous Systems Lab at Stanford University. He obtained a PhD degree from ETH Zurich and M.Sc. and B.Eng. degrees from Tsinghua University. His main research interest is the operation of future mobility systems enabled by connected and automated vehicles (CAVs) and shared mobility.
\end{IEEEbiography}

\vfill
\end{document}